\documentclass[runningheads]{llncs}
\usepackage{graphicx}

\usepackage{caption}
\usepackage{microtype}
\usepackage{booktabs}
\usepackage{multirow} 
\usepackage{marvosym}
\usepackage{cite}
\usepackage{multirow}
\usepackage[section]{placeins}

\newcommand{\refdef}[1]{Definition~\ref{#1}}

\newcommand{\reftheorem}[1]{Theorem~\ref{#1}}
\newcommand{\refeq}[1]{Eq.~\eqref{#1}}

\usepackage{listings}
\usepackage{ifxetex}
\usepackage{fancyhdr}
\usepackage{hyperref}
\usepackage{graphicx}
\usepackage{wrapfig}
\usepackage{amsmath}
\usepackage{amsfonts}
\usepackage{amssymb}
\usepackage{listings}
\usepackage{setspace}
\usepackage{wrapfig}
\usepackage{tablefootnote}
\usepackage{multirow}
\usepackage{booktabs}
\usepackage[section]{placeins}
\usepackage{color}
\usepackage{amsfonts}
\usepackage{ulem}
\usepackage{algorithm}
\usepackage{algorithmicx}
\usepackage{algpseudocode}
\usepackage{graphicx}
\usepackage{float}
\usepackage{dsfont}
\usepackage{cleveref}

\newcommand{\set}[1]{\mathcal{#1}}
\newcommand{\pnorm}[1]{\lVert{#1}\rVert}

\DeclareMathOperator*{\loss}{{\ell}}
\DeclareMathOperator*{\dist}{{d}}

\DeclareMathOperator*{\sign}{sign}

\newcommand{\RN}{\mathbb{R}}

\DeclareMathOperator*{\regularization}{\ensuremath{{\theta}}}
\newcommand{\x}{\ensuremath{\vec{x}}}
\newcommand{\z}{\ensuremath{\vec{z}}}
\newcommand{\y}{\ensuremath{y}}
\newcommand{\xPrime}{\x'}
\newcommand{\yPrime}{\y'}
\newcommand{\xorig}{\ensuremath{\vec{x}_{\text{orig}}}}
\newcommand{\yorig}{\ensuremath{y_{\text{orig}}}}
\newcommand{\setX}{\ensuremath{\set{X}}}
\newcommand{\setY}{\ensuremath{\set{Y}}}
\newcommand{\setI}{\set{I}}
\newcommand{\xcf}{\ensuremath{\vec{x}_{\text{cf}}}}
\newcommand{\xpp}{\ensuremath{\vec{x}_{\text{pp}}}}
\newcommand{\xcfNew}{{\xcf}{_{*}}}

\newcommand{\ycf}{\ensuremath{y'}}
\newcommand{\CF}[2]{\ensuremath{\text{CF}(#1,#2)}}
\newcommand{\deltacf}{\ensuremath{\vec{\delta}}}
\newcommand{\w}{\ensuremath{\vec{w}}}
\newcommand{\wNew}{\ensuremath{\w_{*}}}

\newcommand{\dimsym}{d}
\newcommand{\setD}{\set{D}}
\newcommand{\classifierSet}{\set{H}}
\newcommand{\classifier}{\ensuremath{h}}
\newcommand{\classifierPrime}{\ensuremath{h'}}
\newcommand{\distclassifier}{\ensuremath{\text{d}_{\classifier}}}
\newcommand{\distclassifierPrime}{\ensuremath{\text{d}_{\classifierPrime}}}

\newcommand{\CfDiff}{\ensuremath{\Psi}}
\newcommand{\cfDiff}{\ensuremath{\psi}}

\newcommand{\regression}{\ensuremath{f}}

\begin{document}
%
\title{Contrastive Explanations for Explaining\\ Model Adaptations
\thanks{We gratefully acknowledge funding from the German Federal Ministry of Education and Research (BMBF) through the projects \textit{EML4U} (01IS19080 A) and \textit{TiM} (05M20PBA), and the VW-Foundation for the project \textit{IMPACT} funded in the frame of the funding line \textit{AI and its Implications for Future Society}.}}

%
\titlerunning{Contrastive Explanations for Explaining Model Adaptations}

%
\author{Andr\'e Artelt\inst{1}, Fabian Hinder\inst{1}, Valerie Vaquet\inst{1}, Robert Feldhans\inst{1} \and Barbara Hammer\inst{1}}
\authorrunning{A. Artelt et al.}
%
\institute{CITEC - Cognitive Interaction Technology\\Bielefeld University, 33619 Bielefeld, Germany\\
\email{\{aartelt,fhinder,vvaquet,rfeldhans,bhammer\}@techfak.uni-bielefeld.de}}
\maketitle              
\begin{abstract}
Many decision making systems deployed in the real world are not static - a phenomenon known as model  adaptation takes place over time. The need for transparency and interpretability of AI-based decision models is widely accepted and thus have been worked on extensively. Usually, explanation methods assume a static system that has to be explained. Explaining non-static systems is still an open research question, which poses the challenge how to explain model adaptations.

In this contribution, we propose and (empirically) evaluate a framework for explaining model adaptations by contrastive explanations. We also propose a method for automatically finding regions in data space that are affected by a given model adaptation and thus should be explained.
\end{abstract}
\section{Introduction}
Machine learning (ML) and artificial intelligence (AI) based decision making systems are increasingly affecting our daily life - e.g. predictive policing~\cite{PredictivePolicing} and loan approval~\cite{CreditRiskML,CreditScoresUnfair}. Given the impact of many ML and AI based decision making systems, there is an increasing demand for transparency and interpretability~\cite{DBLP:journals/corr/abs-1906-05684} - the importance of these aspects was also emphasized by legal regulations like the EUs GDPR~\cite{GDPR}. In the context of transparency and interpretability, fairness and other ethical aspects become relevant~\cite{BiasFairnessMLSurvey,FairnessMLSurvey}.

As a consequence, the research community extensively worked on these topics and came up with methods for explaining ML and AI based decision making systems and thus meeting the demands for transparency and interpretability~\cite{ExplainingExplanations,ExplainingBlackboxModelsSurvey,SurveyXai, ExplainableArtificialIntelligence}.
Popular explanations methods~\cite{molnar2019,SurveyXai} are feature relevance/importance methods~\cite{FeatureImportance} and examples based methods~\cite{CaseBasedReasoning}. Instances of example based methods are counterfactual explanations~\cite{counterfactualwachter,CounterfactualReviewChallenges}, influential instances~\cite{InfluentialInstances} and Prototypes \& criticisms~\cite{PrototypesCriticism} - these methods use a set or a single example for explaining the behavior of the system.
Counterfactual and contrastive explanations in general~\cite{molnar2019,counterfactualwachter,CounterfactualReviewChallenges} are popular instances of example based methods for locally explaining decision making systems~\cite{molnar2019} - the reason why these types of explanation are so popular is that because there exists strong evidence that explanations by humans (which they try to mimic) are often counterfactual in nature~\cite{CounterfactualsHumanReasoning}.
While some of these methods are global methods - i.e. explaining the system globally - most of the example based methods are local methods that try to explain the the behavior of the decision making system at a particular instance or in a ``small'' region in data space~\cite{molnar2019,lime2016,LearningGlobalTransparentModelsConsistentWithLocalContrastiveExplanations,botari2020melime}. Further, existing explanation methods can be divided into model agnostic and model specific methods. While model agnostic methods view the decision making system as a black-box and thus do need access to any model internals, model specific methods rely and usually exploit model internal structures and knowledge for computing the explanation. However, distinguishing between model agnostic and model specific methods is not that strict because there exist model specific methods that aim for efficiently computing (initially model agnostic) explanations of specific models~\cite{CounterfactualComputationSurvey}.

The majority of the proposed explanation methods in literature assume fixed models - i.e. explaining the decisions of a fixed decision making system.
However, in practice decision making systems are usually not fixed but (continuously) evolving over time - e.g. the decision making system is adapting or fine tuned on new data~\cite{LifeLongLearningNeuralNetworksReview}. In this context it becomes relevant to explain the changes of the decision making system\footnote{E.g. The authors of~\cite{PhilosophyAlgorithmicRecourse} discuss the problem that explanations of a changing classifier can become invalid (i.e. expire) after some time and thus pose a major problem in algorithmic recourse.} - in particular in the context of Human-Centered AI (HCAI) which, besides explainability, is another important building block\footnote{Some people even argue that explainability and transparency are an essential part of HCAI~\cite{ComputerSaysNo,10.1145/3419764}} in ethical AI~\cite{wortmanvaughan2021a}. HCAI allows the human being (the people) to ``rule'' AI systems instead of being ``discriminated'' or ``cheated'' by AI. Given the complexity of many modern ML or AI systems (e.g. Deep Neural Networks), it is usually difficult for a human to understand the decision making system or the impact of some adaptation or changes applied to a given system.
Yet, understanding the impact of changing a system in a given way is crucial for rejecting system changes that violates some (ethical) guidelines or (legal) constraints.

For example if we consider the scenario of a (non-trivial) loan approval system that automatically accepts or rejects loan applications - i.e. we assume that the decision making process of this system is highly complicated and difficult to inspect from the outside (e.g. it might be a Deep Neural Network):
\textit{We might encounter a situation in which a loan application was rejected with the argument of a low income and a bad payback rate in the past - which perfectly meets the bank internal guidelines for accepting or rejecting a loan. Next, we adapt the loan approval system on new data - we assume that we got new data for fine tuning the system - but the we assume that the guidelines or policies for accepting or rejecting did not change.
But after this adaptation, it turns out that the same application that was rejected under the ``old'' system (before the adaptation), it is now accepted by the new system - that is we assumes that this changed behavior violates the risk-guidelines of the bank because it exposes the bank to an unnecessarily higher risk of loosing money.}
This is an example in which case we would like to reject the given model adaptation because this adaptation would lead to a system that violates some predefined rules. Since in practice we do not always have a detailed policy available\footnote{Otherwise there would be no or very limited need for using some ML or AI system that learns such a policy from data.}, we need a mechanism that makes the impact of model adaptations/changes transparent so that it can be ``approved'' by a human\footnote{Ideally the explanation of the adaptations/changes are simple enough to be understood by lay persons instead of only be accessible to ML or AI experts.}.

Although there exist general overview work that is aware of the challenge of explaining changing ML systems~\cite{CounterfactualReviewChallenges}, how to exactly do this is still an open research question which we aim to address in this contribution.
In this work, we propose a framework that uses contrastive explanations for explaining model adaptation - i.e. we argue that inspecting the changes/differences of contrastive explanations is a reasonable proxy for explaining model adaptations.
More precisely, our contributions are:
\begin{itemize}
    \item We propose to compare contrasting explanations for locally explaining model adaptations.
    \item We propose a method for finding relevant and interesting regions in data space which are affected by a given model adaptation and thus should be explained to the user.
    \item We propose persistent local explanations for regularizing the model adaptation towards models with a specific behaviour.
\end{itemize}
The remainder of this work is structured as follows: After briefly reviewing literature and taking a look at the foundations like contrasting explanations (section~\ref{sec:foundations:contrastiveexplanations}) and model adaptations (section~\ref{sec:foundations:modeladaptations}) we introduce and describe our proposal for using contrastive explanations for locally explaining model adaptations in section~\ref{sec:modeling}. In this context, we then study counterfactual explanations as a specific instance of contrastive explanations in section~\ref{sec:diffexplanations:counterfactuals} - in particular we study counterfactuals for linear models (see section~\ref{sec:diffexplanations:linearapproxmodels}) and propose a method for finding relevant and interesting and relevant regions in data space (see section~\ref{sec:finding_relevant_regions_in_dataspace}).
In section~\ref{sec:constrained_model_adaptations}, we introduce our idea of persistent contrasting explanations - we consider different types of persistent explanations constraints and study how to add them to the model adaptation optimization problem.
Finally, we empirically evaluate our proposed methods in section~\ref{sec:experiments} and close our work with a  summary and discussion in section~\ref{sec:conclusion}.

Due to space constraints and for the purpose of better readability, we include all proofs and derivations in appendix~\ref{sec:appendix}.

\paragraph*{Related work} 
While (concept) drift as well as transparency (i.e. methods for explaining decision making systems) have been extensively studied separately, the combination of both have received much less attention so far.

Counterfactual explanations are a popular instance of example based explanation methods but all existing methods so far assume that the underlying model which is explained does not change over time - a strategy for counterfactual explanations of changing/drifting models is still missing~\cite{CounterfactualReviewChallenges}.

A method called ``counterfactual metrics''~\cite{artelt2021neurocomputing} can be used for explaining drifting feature relevances of metric based models. In contrast to a counterfactual explanation, it focuses on feature relevance rather than change of counterfactual examples. The authors of~\cite{artelt2021neurocomputing} consider a scenario in which a metric based model is adapted to drifting feature relevances and the resulting model adaptation is explained by the the changes made to the distance metric which they call ``counterfactual metrics''.

In~\cite{hinder2020counterfactual}, contrastive explanations (in particular counterfactual explanations) are used for explaining concept drift. For this purpose a classifier is constructed which tries to separate two batches of data that are assumed to be affected by concept drift - the concept drift is explained by using contrastive explanations that contrast a sample from one class (i.e. batch) to the other class/batch under the trained classifier. The authors also propose a method for finding interesting and relevant samples which they call ``characteristic samples'' that are affected by the concept drift and thus promising candidates for illustrating and explaining the present drift.

\section{Foundations}\label{sec:foundations}
\subsection{Model Adaptions}\label{sec:foundations:modeladaptations}
We assume a model adaptation scenario in which we are given a prediction function (also called model) $\classifier(\cdot)$ and a set of (new) labeled data points $\setD$. Adapting the model $\classifier(\cdot)$ to the data $\setD$ means that we want to find a model $\classifierPrime(\cdot)$ which is as similar as possible to the original model $\classifier(\cdot)$ while performing well on labeling the (new) samples in $\setD$~\cite{LifeLongLearningNeuralNetworksReview}. Model adaptation can be formalized as an optimization problem~\cite{LifeLongLearningNeuralNetworksReview} as stated in~\refdef{def:modeladaptation}.
\begin{definition}[Model adaptation]\label{def:modeladaptation}
Let $\classifier:\setX\to\setY$, $\classifier\in\classifierSet$ be a prediction function (also called model) and $\setD=\{(\x_i,\y_i)\in\setX\times\setY\}$ a set of (new) labeled data points.
Adapting the model $\classifier(\cdot)$ to the data $\setD$ is formalized as the following optimization problem:
\begin{equation}\label{eq:modeladaptation:opt}
    \underset{\classifierPrime\,\in\,\classifierSet}{\arg\min}\,\regularization(\classifier, \classifierPrime) \quad \text{s.t. }\classifierPrime(\x_i)\approx\y_i \quad \forall\,(\x_i,\y_i) \in \setD
\end{equation}
where $\regularization:\classifierSet\times\classifierSet\to\RN_{+}$ denotes a regularization that measures the  similarity between two given models
\footnote{In case of a parameterized model, one possible regularization measures the difference in the parameters.}
and $\approx$ refers to a suitable prediction error (e.g. zero-one loss or squared error) which is minimized by $\classifierPrime(\cdot)$.
\end{definition}
Note that for large $\setD$, e.g.\ caused by abrupt concept drift  \cite{gama2014survey}, one could completely retrain $\classifier(\cdot)$ and abandon the requirement of closeness.
In such situations it is still interesting to explain the difference of $\classifierPrime(\cdot)$ and  $\classifier(\cdot)$.
%

\subsection{Contrastive Explanations}\label{sec:foundations:contrastiveexplanations}
Counterfactual explanations are a popular instance of contrastive explanations. A counterfactual explanations - often just called counterfactual or pertinent positive by some authors~\cite{ContrastiveExplanationsWithPertinentNegatives} - states a change to some features of a given input such that the resulting data point (called counterfactual) has a different (specified) prediction than the original input. The rational is considered to be intuitive, human-friendly and useful because it tells the user which minimum  changes can lead to a desired outcome~\cite{molnar2019,counterfactualwachter}. Formally, a (closest) counterfactual can be defined as follows:
\begin{definition}[(Closest) Counterfactual Explanation~\cite{counterfactualwachter}]\label{def:counterfactual}
Assume a prediction function $\classifier:\RN^\dimsym \to \setY$ is given. Computing a counterfactual $\xcf \in \RN^\dimsym$ for a given input $\x \in \RN^\dimsym$ is phrased as an optimization problem:
\begin{equation}\label{eq:counterfactualoptproblem}
\underset{\xcf \,\in\, \RN^\dimsym}{\arg\min}\; \loss\big(\classifier(\xcf), \ycf\big) + C \cdot \regularization(\xcf, \x)
\end{equation}
where $\loss(\cdot)$ denotes the loss function,  $\ycf$ the requested prediction, and  $\regularization(\cdot)$  a penalty term for deviations of $\xcf$ from the original input $\x$. $C>0$ denotes the regularization strength.
\end{definition}
\begin{remark}
In the following we assume a binary classification problem. In this case we denote a (closest) counterfactual $\xcf$ according to~\refdef{def:counterfactual} of a given sample $\x$ under a prediction function $\classifier(\cdot)$ as $
\xcf=\CF{\x}{\classifier}$ as the desired target is uniquely determined.
\end{remark}
The authors of~\cite{ContrastiveExplanationsWithPertinentNegatives} define a contrastive explanations consisting of two parts: a pertinent negative (counterfactual) and a pertinent positive. A pertinent positive~\cite{ContrastiveExplanationsWithPertinentNegatives,ContrastiveExplanationsModelAgnostic,EfficientComputationContrastiveExplanations} describes a minimal set of features that are already sufficient for the given prediction. As already mentioned, pertinent positives are usually considered as a part or addition of contrastive explanations and are assumed to provide additional insights why the model took a particular decision~\cite{ContrastiveExplanationsWithPertinentNegatives}.

A pertinent positive of a sample $\xorig$ describes a minimal set of features $\setI$ (of this particular sample $\xorig$) that are sufficient for getting the same prediction - these features are also called ``turned on'' features and all other features are called ``turned off'' meaning that they are set to zero or any other specified default value. One could either require that all ``turned on'' features are equal to their original values in $\xorig$ or one could relax this and only require that they are close to their original values in $\xorig$.
The computation of a pertinent positive (also called sparsest pertinent positive) can be phrased as the following multi-objective optimization problem~\cite{EfficientComputationContrastiveExplanations}:
\begin{subequations}\label{eq:pertinentpositive:modelling}
\begin{align}
    & \underset{\deltacf\,\in\,\RN^{\dimsym}}{\min}\; \Big|\left[\xorig - \xcf\right]_{\setI}\Big| \quad \text{where } \xcf = \xorig - \deltacf\\
    & \underset{\setI}{\min}\; \left|\setI\right|\\
    & \text{s.t.} \quad \classifier(\xcf) = \yorig
\end{align}
\end{subequations}
where $[\cdot]_{\setI}$ denotes the selection operator on the set $\setI$ and $\setI$ denotes the set of all ``turned on'' features.\footnote{The selection operator returns a vector, whereby it only selects a subset of indices from the original vector as specified in the set $\setI$.} $\setI$ is defined as follows:
\begin{equation}\label{eq:setI}
\setI = \Big\{i:\,\big|(\xcf)_i\big|>\epsilon\Big\}
\end{equation}
where $\epsilon\in\RN_{+}$ denotes a tolerance threshold at which we consider a feature ``to be turned on'' - e.g. a strict choice would be $\epsilon=0$.

Since~\eqref{eq:pertinentpositive:modelling} is difficult to solve - a number of different methods for efficiently computing (approximate\footnote{The approximation comes from giving up closeness - many methods successfully compute pertinent positives but can not guarantee that they are globally optimal.}) pertinent positives have been proposed~\cite{ContrastiveExplanationsWithPertinentNegatives,ContrastiveExplanationsModelAgnostic,EfficientComputationContrastiveExplanations}.

\section{Explaining Model Adaptations}\label{sec:modeling}
An obvious approach for explaining a model adaptation would be to explain and communicate the regions in data space where the prediction of the new and the old model are different - i.e. $\{\x\in\setX \mid\classifier(\x)\neq\classifierPrime(\x)\}$. However, in particular for incremental model adaptations, this set might be small and its characterization not meaningful. Hence, instead of finding these samples, where the two models compute different predictions\footnote{Which of course is an obvious and reasonable starting point for explaining the difference between two models.}, we aim for an explanation of their learned generalization rules. Because of the constraint in~\refeq{eq:modeladaptation:opt}, it is likely the case that both models compute the same prediction on all samples in a given test. However, the reason for these predictions can be arbitrarily different (depending on the regularization $\regularization(\cdot)$) - i.e. the internal prediction rules of both models are different. We think that explaining and thus understanding how and where the reasons and rules for predictions differ are much more useful than just inspecting points where the two models disagree - in particular when it comes to understand and judging decision making systems in the context of human centered AI.

In the following,
we propose that a contrastive explanation can serve as a proxy for
the model generalization at a given point; hence a comparison of
the possibly different contrastive explanations of two  models at a given point 
can be considered as an explanation of how the different underlying principles based on which the models propose a decision. 
%
As it is common practice, we thereby look at local differences, since a global comparison might be too complex and not easily understood by a human 
~\cite{molnar2019,CounterfactualReviewChallenges,CounterfactualsHumanReasoning}.
Furthermore, it might also be easier to add constraints on specific samples instead of constraints on the global decision boundary to the model adaptation problem~\refeq{eq:modeladaptation:opt} in~\refdef{def:modeladaptation}.
The computation of such differences of explanations is tractable as long as the computation of the constrastive explanation under a single model itself is tractable - i.e. no additional computational complexity is introduced when using this approach for explaining model adaptations.

\subsection{Modeling}
We define this type of explanation as follows:
\begin{definition}[Explaining Model Differences]\label{def:diffexplanations}
We assume that we are given a set of labeled data points $\setD^{*}=\{(\x^{*}_i, \y^{*}_i)\in\setX\times\setY\}$ whose labels are correctly predicted by both models $\classifier:\setX\to\setY$ and $\classifierPrime:\setX\to\setY$.

For every $(\x^{*}_i, \y^{*}_i)\in\setD^{*}$, let $\deltacf^{i}_{\classifier}\in\setX$ be a contrastive explanation under $\classifier(\cdot)$ and  $\deltacf^{i}_{\classifierPrime}\in\setX$ under the new model $\classifierPrime(\cdot)$. The explanation of the model differences at point $(\x^{*}_i, \y^{*}_i)$ and its magnitude is then given by the comparison of both explanations:
\begin{equation}
 \cfDiff(\deltacf^{i}_{\classifier}, \deltacf^{i}_{\classifierPrime}) \mbox{\ and\ }
    \CfDiff(\deltacf^{i}_{\classifier}, \deltacf^{i}_{\classifierPrime})
\end{equation}
where $\cfDiff(\cdot)$ denotes a suitable operator which can compare the information contained in two given explanations and
$\CfDiff(\cdot)$ denotes a real-valued distance measure judging the difference of explanations.
\end{definition}
\begin{remark}
Note that the explanation as defined in~\refdef{def:diffexplanations} can be more generally applied to compare two classifiers
$\classifierPrime(\cdot)$ and $\classifier(\cdot)$ w.r.t.\ given input locations, albeit $\classifierPrime(\cdot)$ does not constitute an adaptation of $\classifier(\cdot)$. For simplicity, we assume uniqueness of contrastive explanations in the definition, either by design such as given for linear models or 
by suitable algorithmic tie breaks.
\end{remark}
The concrete form of the explanation heavily depends on the comparison function $\cfDiff(\cdot)$ and $\CfDiff(\cdot)$ - this allows us to take specific use-cases and target groups into account.
In this work we assume $\setX=\RN^\dimsym$ and $\cfDiff(\cdot)$
is given by the component-wise absolute value 
$|(\deltacf^{i}_{\classifier})_l-(\deltacf^{i}_{\classifierPrime})_l|_l$, and we
consider two possible realizations of $\CfDiff(\cdot)$:
\paragraph*{Euclidean similarity}
An obvious measurement of the difference of two explanations is to compute the Euclidean distance between them:
\begin{equation}\label{eq:cfdiff:euclid}
    \CfDiff(\deltacf_{\classifier}, \deltacf_{\classifierPrime}) = \pnorm{\deltacf_{\classifier} -  \deltacf_{\classifierPrime}}_2
\end{equation}
where on could also use a different $p$-norm (e.g. $p=1$).

\paragraph*{Cosine similarity}
We can also measure the difference between two explanations by the cosine of their respective angle:
\begin{equation}\label{eq:cfdiff:cosine}
    \CfDiff(\deltacf_{\classifier}, \deltacf_{\classifierPrime}) = \cos\left(\angle \deltacf_{\classifier}, \deltacf_{\classifierPrime}\right)
\end{equation}

\begin{remark}
Note that considering the angle, instead of the actual distance, has the  advantage that it is scale invariant - i.e. it is more interesting if different features have to be changed, rather than the same features have to be changed slightly more.

Since the image of the cosine is limited to $[-1,1]$, we can directly compare the values of different samples - whereas the general norm is only capable of comparing nearby points with each another as it contains information regarding the local topological structure.
\end{remark}

\section{Counterfactual Explanations for Explaining Model Adaptations}\label{sec:diffexplanations:counterfactuals}
Counterfactual explanations~\refdef{def:counterfactual} are a popular instance of contrastive explanations (section~\ref{sec:foundations:contrastiveexplanations}). In the following, we study counterfactual explanations for explaining model adaptations as proposed in~\refdef{def:diffexplanations}.
We first relate  
the difference between two linear classifiers to their counterfactuals and, vice versa,  the change of counterfactuals to model change.
Finally, we propose a method for finding relevant regions and samples for comparing counterfactual explanations.

\subsection{Counterfactuals for Linear Models}\label{sec:diffexplanations:linearapproxmodels}
First, we highlight the possibility to relate the similarity of two linear models at a given point to their counterfactuals.
We consider a linear binary classifier $\classifier: \RN^\dimsym \to \{-1,1\}$:
\begin{equation}\label{eq:linearclassifier}
    \classifier(\x) = \sign(\w^\top\x)
\end{equation}
and assume w.l.o.g. that the weight vector $\w\in\RN^\dimsym$ has unit length, $\pnorm{\w}_2=1$.
We assume for an adaptation $\classifierPrime(\x) = \sign(\wNew^\top\x)$ with unit weight vector $\wNew$

In~\reftheorem{theorem:linearmodels:cosine} we state how to use counterfactuals for approximately computing the local cosine similarity between two models - we interpret this as evidence for the usefulness of counterfactual explanations for measuring the difference between two given models.
\begin{theorem}[Cosine Similarity of Linear Models]\label{theorem:linearmodels:cosine}
Let $\classifier(\cdot)$ and $\classifierPrime(\cdot)$ be two linear models, and $\x$ a data point. 
Let $\xcf=\CF{\x}{\classifier}$ and $\xcfNew=\CF{\x}{\classifierPrime}$ be the closest counterfactual of a data point $\x\in\RN^\dimsym$ under the original model resp. the adapted model $\classifierPrime(\cdot)$. 
Then 
\begin{equation}\label{eq:generalidea:locallinearapprox:diff}
   \cos(\angle\w,\wNew)= \frac{\xcf^\top\xcfNew + \x^\top\x - \xcf^\top\x - \xcfNew^\top\x}{\sqrt{\left(\xcf^\top\xcf + \x^\top\x - 2\xcf^\top\x\right)\left(\xcfNew^\top\xcfNew + \x^\top\x - 2\xcfNew^\top\x\right)}}
\end{equation}
\end{theorem}
Since every (possibly nonlinear) model can locally be approximated linearly, this result also indicates the relevance of counterfactuals to characterize local differences of two models.

Conversely, it is possible to limit the difference of counterfactual explanations by the difference of classifiers as follows:
\begin{theorem}[Change of a Closest Counterfactual Explanation]\label{theorem:counterfactualdrift:linearmodel}
Let $\classifier(\cdot)$ be a binary linear classifier~\refeq{eq:linearclassifier} and $\classifierPrime(\cdot)$ be its adaptation.
Then, the difference between the two closest counterfactuals of an arbitrary sample $(\x,\y)\in\RN^\dimsym\times\{\-1,1\}$ can be bounded as:
\begin{equation}\label{eq:linearmodel:bound:counterfactual}
    \pnorm{\CF{\x}{\classifier} - \CF{\x}{\classifierPrime}}_2 \leq \sqrt{8}\pnorm{\x}_2\big(1 - \cos(\angle\w,\wNew)\big)^{1/2}
\end{equation}
\end{theorem}

\subsection{Finding Relevant Regions in Data Space}\label{sec:finding_relevant_regions_in_dataspace}
The task of sample based model comparison obviously requires the selection of feasible samples, as the amount of samples is usually to large to be dealt with by hand. Thus, we need to formalize a notion of characteristic samples in the context of model change to perform this sub-task automatically. In this section, we aim to formalize this notion and give a number of possible choices and respective approximation regarding this problem.

The idea is to provide an interest function, i.e. a function that marks the regions of the data space that are of interest for our consideration - we could use such a function for automatically finding interesting samples by applying it to a set of points to get a ranking or optimizing over the function for coming up with interesting samples. This function $i(\cdot)$ should have certain properties:
\begin{enumerate}
    \item For every pair of fixed models $\classifier,\classifierPrime \in \classifierSet$ it maps every point $x \in \setX$ in the data space to a non-negative number - i.e. $i : \setX \times (\classifierSet \times \classifierSet) \to \RN_{+}$.
    \item It should be continuous with respect to the classifiers and in particular $i(x,\classifier,\classifierPrime) = 0$ for all $x$ if and only if $\classifier= \pm \classifierPrime$. 
    \item Points that are ``more interesting'' should take on higher values.
    \item Regions where the classifiers coincide are not of interest.
\end{enumerate}

The last two properties are basically a localized version of the second in the sense that it forces $i(\cdot)$ to turn properties of the decision boundary (which are global) into local, i.e. point wise, properties. It is possible to make the properties 3 and 4 rigorous, but this would require an inadequate amount of theory. 

An obvious definition of $i(\cdot)$ is to directly use the explanation~\refdef{def:diffexplanations} itself together with a difference measurement $\CfDiff(\cdot)$ as stated in~\refeq{eq:cfdiff:euclid} and~\refeq{eq:cfdiff:cosine}:
\begin{equation}\label{eq:interestfunction}
    i(\x,\classifier,\classifierPrime) = \CfDiff\big(\CF{\x}{\classifier}, \CF{\x}{\classifierPrime}\big)
\end{equation}
Then it is easy to see that the four properties are fulfilled, if we assume that $\CfDiff(\cdot)$ and is chosen correctly:

The first property follows from the definition of dissimilarity functions and so does the second. The fourth property follows from the fact that if the classifiers intrinsically perform the same computations then the counterfactuals are the same and hence $i(\cdot) = 0$; on the other hand, if the classifiers (intrinsically) perform different computations (thought the overall output may be the same) then the counterfactuals are different and hence the $i(\cdot) \neq 0$. In a comparable way the third property is reflected by the idea that obtaining counterfactuals is faithful to the computations in the sense that slight resp. very different computation will lead to slight resp. very different counterfactuals.

Besides the Euclidean distance~\refeq{eq:cfdiff:euclid}, the cosine similarity~\refeq{eq:cfdiff:cosine} is a potential choice for comparing two counterfactuals in $\CfDiff(\cdot)$.
Since the cosine always takes values between $-1$ and $1$, we scale it to a positive codomain:
\begin{equation}\label{eq:cfdiff:cosine2}
    \CfDiff\big(\CF{\x}{\classifier}, \CF{\x}{\classifierPrime}\big) = 2 - \cos\big(\angle \CF{\x}{\classifier}, \CF{\x}{\classifierPrime}\big)
\end{equation}
However, $\CfDiff(\cdot)$ as defined in~\refeq{eq:cfdiff:cosine2} is discontinuous if we approach the decision boundary of one of the classifiers.
This problem can be resolved by using an relaxed version of~\refeq{eq:cfdiff:cosine2}:
\begin{equation}
    \CfDiff\big(\CF{\x}{\classifier}, \CF{\x}{\classifierPrime}\big) = 2-\frac{\langle \CF{\x}{\classifier}-\x,\CF{\x}{\classifierPrime}-\x \rangle}{\pnorm{\CF{\x}{\classifier}-\x}_2 \pnorm{\CF{\x}{\classifierPrime}-\x}_2 + \varepsilon}
\end{equation}
for some small $\varepsilon > 0$. In this case the samples on the decision boundary are marked as not interesting, which fits the finding that the counterfacutals for those samples basically coincide with the samples them self and therefore do not provide any (additional) information.

\paragraph{An approximation for gradient based models}
While the definition of the interest function~\refeq{eq:interestfunction} perfectly captures our goal of identifying interesting samples, it can be computational difficult to compute. In particular the computation of (closest) counterfactual explanations can be computational expensive and ``challenging'' for many models~\cite{CounterfactualComputationSurvey} - this becomes a major issue when optimizing, that is searching for local maxima of $i(\cdot)$. It is hence of importance to find a surrogate for the counterfactual $\CF{\cdot}{\cdot}$ that allows for fast and easy computation.

In these cases, an efficient approximation is possible,
provided the classifier $\classifier(\cdot)$ is induced by a differentiable function $\regression(\cdot)$ in the form  $\classifier(\x) = \sign\left(\regression(\x)\right)$.
Then the gradient of $\regression(\cdot)$ enables an approximation of the counterfactual in the following form:
\begin{equation}\label{eq:interestfunction:cfsurrogate}
    \CF{\x}{\classifier} = \x - \eta \classifier(\x) \nabla_{\x} \regression(\x)
\end{equation}
for a sufficient $\eta > 0$.
In this case~\refeq{eq:interestfunction:cfsurrogate}, the cosine similarity approach~\refeq{eq:cfdiff:cosine2} works particularly well because it is invariant with respect to the choice of $\eta$ - i.e. $\eta$ can be ignored and we only need the gradient.
Another benefit of this choice is that, under some smoothness assumptions regarding the classifier, admits simple geometric interpretations of the obtained values as the gradient always point towards the closes point on the decision boundary. This way it (locally) reduces the interpretation to linear classifiers for which counterfactual explanations are well understood~\cite{CounterfactualComputationSurvey}.

In the remainder of this work, we use the gradient approximation together with the cosine similarity for computing the ``usefulnes'' of given samples for comparing their counterfactual explanations.
\section{Constrained Model Adaptation for Persistent Explanations}\label{sec:constrained_model_adaptations}
In the previous sections we proposed and studied the idea of comparing contrastive (in particular counterfactual) explanations for explaining model adaptations. In the experiments (see section~\ref{sec:experiments}) we observe this method is indeed able to detected and explain less obvious  and potential problematic changes of the internal decision rules of adapted models.

In this context of explaining model adaptations, Human-Centered AI comes into play when the user rejects the computed model adaptation based on the explanations. For instance it might happen, that the local explanation under the old model $\classifier(\cdot)$ was accepted, but the new local explanation under the new model $\classifierPrime(\cdot)$ violates some rules or laws - see the introduction for an example.
In such a case, we want to constrained the model adaptation~\refdef{def:modeladaptation} such that (some) local explanations remain the same or valid under the new model - i.e. making some local explanations persistent - and to push the new model $\classifierPrime(\cdot)$ towards globally accepted behavior by making use of such local constraints.

\subsection{Persistent Local Explanations}\label{sec:general:persitent_local_explanations}
For the purpose of ``freezing'' a local explanation in the form of a contrastive explanation - i.e. making it persistent -, we propose the following (informal) requirements:
\begin{itemize}
    \item Distance to the decision boundary must be within a given interval.
    \item Counterfactual explanation must be still (in-)valid.
    \item Pertinent positive must be still (in-)valid.
\end{itemize}
Aiming for persistent contrastive explanations, we have to augment the original optimization problem~\refeq{eq:modeladaptation:opt} from~\refdef{def:modeladaptation} for adapting a given model $\classifier(\cdot)$ as follows:
\begin{subequations}\label{eq:modeladaptation_constrained:opt}
\begin{align}
    &\underset{\classifierPrime\,\in\,\classifierSet}{\arg\min}\,\regularization(\classifier, \classifierPrime) \label{eq:modeladaptation_constrained:regularization} \\
    &\text{s.t. }\classifierPrime(\x_i)\approx\y_i \quad \forall\,(\x_i,\y_i) \in \setD \label{eq:modeladaptation_constrained:constraint:correct_classification} \\
    &\quad\;\;\text{Some local contrastive explanations under $\classifier$ are still true under $\classifierPrime$} \label{eq:modeladaptation_constrained:constraint:persistent_explanation}
\end{align}
\end{subequations}
where the additional (informal) constraint~\refeq{eq:modeladaptation_constrained:constraint:persistent_explanation} is the only difference to the original optimization problem~\refeq{eq:modeladaptation:opt}.

Next, we study how we can formalize~\refeq{eq:modeladaptation_constrained:constraint:persistent_explanation} and thus how to solve~\refeq{eq:modeladaptation_constrained:opt} for different models and different explanation constraints.

\subsection{Modeling}\label{sec:constrained_model_adaptations:modeling}
We always assume that we are given a labeled sample $(\xorig,\yorig)\in\setX\times\setY$ that is correctly labeled by the old model $\classifier(\cdot)$ as well as the new model $\classifierPrime(\cdot)$ - i.e. $\classifier(\xorig)=\classifierPrime(\xorig)=\yorig$. In the subsequent section, we study how to write constraint~\refeq{eq:modeladaptation_constrained:constraint:persistent_explanation} in~\refeq{eq:modeladaptation_constrained:opt} for the different requirements/constraints as listed in section~\ref{sec:general:persitent_local_explanations} - it turns out that we can often write the constraints (at least after a reasonable relaxation) as additional labeled samples which enable a straight forward incorporation into many model adaption procedures (see section~\ref{sec:constrained_model_adaptations:implementations} for details):
\begin{equation}
    \classifierPrime(\xPrime)=\yPrime
\end{equation}

\subsubsection{Persistent Distance to Decision Boundary}
In case of a classifier, one might require that the distance to the decision boundary \distclassifier(\xorig) is not larger than some fixed $\lambda\in\RN_{+}$.
Applying this to our model adaption setting, we get the following constraint:
\begin{equation}
    \distclassifierPrime(\xorig) \leq \lambda
\end{equation}
However, because reasoning over distances to decision boundaries might be a too complicated and often difficult to formalize as a computational tractable constraint, one might instead require that all samples that are ``close'' or ``similar'' to $\xorig$ must be have the same prediction $\yorig$:
\begin{equation}\label{eq:constraint:similarpoints}
    \classifier(\x)=\yorig \quad \forall\,\x\in\mathcal{E}(\xorig, \setX)
\end{equation}
where we defined the set of all ``similar''/''cose'' points as follows:
\begin{equation}
    \mathcal{E}(\xorig, \setX) = \big\{\x \in \setX \mid \dist(\x,\xorig)\leq\lambda\big\}
\end{equation}
where $\dist:\setX\times\setX\to\RN_{+}$ denotes an arbitrary similarity/closeness measure - e.g. in case of real valued features the $p$-norm might be a popular choice.
If $\mathcal{E}(\xorig, \setX)$ contains a ``small'' number of elements only, then~\refeq{eq:constraint:similarpoints} is computational tractable and can be added as a set of constraints to the optimization problem~\refeq{eq:modeladaptation_constrained:opt}. However, in case of real valued features where we use the $p$-norm (e.g. $p=1$ or $p=2$) as a closeness/similarity measure, we get the following constraint:
\begin{equation}\label{eq:constraint:adversarial_robustness}
    \classifier(\x)=\yorig \quad \forall\,\x\in\setX: \; \pnorm{\x - \xorig}_{p} \leq \lambda
\end{equation}
Note that constraints of the form of~\refeq{eq:constraint:adversarial_robustness} are well known and studied in adversarial robustness literature~\cite{carlini2019evaluating} - these constraints reduce the problem to a training an locally adversarial robust model $\classifierPrime(\cdot)$.

Further relaxing the idea of a persistent distance to the decision boundary might lead to requirements where a set of features is increased or decreased such that the original prediction remains the same. For instance one might have a set of $\z_j\in\RN^\dimsym$ which must not change the prediction if added to the original sample $\xorig$, yielding the following constraint:
\begin{equation}\label{eq:constraint:robustness}
    \classifierPrime(\xorig + \z_j) = \yorig \quad \forall\,j
\end{equation}

\subsubsection{Persistent Counterfactual Explanation}
Recall that in a counterfactual explanation, we add a perturbation $\z\in\RN^\dimsym$ to the data point $\xorig$ which results in a (specified) prediction different from $\yorig$:
\begin{equation}
    \classifier(\xorig + \z) = \classifier(\xcf) = \ycf \neq \yorig
\end{equation}
where we defined $\xcf=\xorig + \z$.

Requiring that the same counterfactual explanations holds for the adapted model $\classifierPrime(\cdot)$, yields the following constraint:
\begin{equation}\label{eq:constraint:persitent_cf}
    \classifierPrime(\xorig + \z) = \classifierPrime(\xcf) = \ycf
\end{equation}
Note that with constraint~\refeq{eq:constraint:persitent_cf} alone, we can not guarantee that $\xcf$ will be the closest counterfactual of $\xorig$ under $\classifierPrime(\cdot)$ - although it is guaranteed to be a valid counterfactual explanation. However, we think that computing the closest counterfactual is not that important because the closest counterfactual is very often an adversarial which might not be that useful for explanations~\cite{molnar2019,PlausibleCounterfactuals} and for sufficiently complex models, computing the closest counterfactual becomes computational difficult~\cite{CounterfactualComputationSurvey}. Furthermore, closeness becomes even less important when dealing with plausible counterfactuals which are usual not the closest ones~\cite{PlausibleCounterfactuals} - if $\xcf$ is a plausible counterfactual under $\classifier(\cdot)$ one would expect that it is also plausible under $\classifierPrime(\cdot)$ because the data manifold itself is not expected to change that much.

\subsubsection{Persistent Pertinent Positive}
Recall that a pertinent positive $\xpp\in\RN^\dimsym$ describes a sparse sample where all non-zero feature values are as close as possible to the feature values of the original sample $\xorig$ and the prediction is still the same:
\begin{equation}
    \classifier(\xorig) = \classifier(\xpp) = \yorig
\end{equation}
Requiring that $\xpp$ is still a pertinent positive of $\xorig$ under the adapted model $\classifierPrime(\cdot)$, yields the following constraint:
\begin{equation}\label{eq:constraint:persitent_pp}
    \classifierPrime(\xpp) = \yorig
\end{equation}
Similar to the case of persistent counterfactual explanations,~\refeq{eq:constraint:persitent_pp} does not guarantee that $\xpp$ is the sparsest or closest pertinent positive of $\xorig$ under $\classifierPrime(\cdot)$ - it could happen that there exists an even sparser or closer pertinent positive of $\xorig$ under $\classifierPrime(\cdot)$ which was invalid under the old model $\classifier(\cdot)$. However, it is guaranteed that $\xpp$ is a sparse pertinent positive of $\xorig$ under $\classifierPrime(\cdot)$ which we consider to be sufficient for practical purposes, in particular if taking into account the computational difficulties of computing a pertinent positive - as stated in~\cite{EfficientComputationContrastiveExplanations}, computing a pertinent positive (even of ``classic'' ML models) is not that easy.

\subsection{Model specific Implementation}\label{sec:constrained_model_adaptations:implementations}
We consider a scenario where we have a sample wise loss function\footnote{E.g. smth. like the squared error or negative log-likelihood.} ${\loss}_{\classifierPrime}:\setX\times\setY\to\RN$ that penalizes prediction errors and a set of (new) labeled data points $\setD=\{(\x_i,\y_i)\in\setX\times\setY\}$ to which we wish to adapt our original model $\classifier(\cdot)$ - we rewrite the model adaptation optimization problem~\refeq{eq:modeladaptation:opt} as follows:
\begin{equation}\label{eq:modeladaptation_general:opt}
    \underset{\classifierPrime\,\in\,\classifierSet}{\arg\min}\,\regularization(\classifier, \classifierPrime) + C\sum_i {\loss}_{\classifierPrime}(\x_i,\y_i)
\end{equation}
where the hyperparameter $C\in\RN_{+}$ allows us to balance between closeness and correct predictions.

Next, we assume that we have a bunch of persistence constraints $\setD^{*}=\{(\xPrime_j, \yPrime_j)\in\setX\times\setY\}$ of the form $\classifierPrime(\xPrime_j)=\yPrime_j$ as discussed in the previous section~\ref{sec:constrained_model_adaptations:modeling}. Considering these constraints, we rewrite the constrained model adaptation optimization problem~\refeq{eq:modeladaptation_constrained:opt} as follows:
\begin{equation}\label{eq:modeladaptation_general:constrained:opt}
        \underset{\classifierPrime\,\in\,\classifierSet}{\arg\min}\,\regularization(\classifier, \classifierPrime) + C\sum_i {\loss}_{\classifierPrime}(\x_i,\y_i) +  C'\sum_j {\loss}_{\classifierPrime}(\xPrime_j,\yPrime_j)
\end{equation}
where we introduce a hyperparameter $C'\in\RN_{+}$ that denotes a regularization strength which, similar to the hyperparameter $C$, helps us enforcing satisfaction of the additional persistence constraints - encoded as~\refeq{eq:modeladaptation_constrained:constraint:persistent_explanation} in the original informal modelling~\refeq{eq:modeladaptation_constrained:opt}.

Assuming a parameterized model, we can use any black-box optimization method (like Downhill-Simplex) or a gradient-based method if~\refeq{eq:modeladaptation_general:constrained:opt} happens to be fully differentiable with respect to the model parameters. However, such methods usually come without any guarantees and are highly sensitive to the solver and the chosen hyperparameters $C$ and $C'$. Therefore, one would be advised to use exploit model specific structures for efficiently solving~\refeq{eq:modeladaptation_general:constrained:opt} - e.g. write~~\refeq{eq:modeladaptation_general:constrained:opt} in constrained form and turn it into a convex program.

\section{Experiments}\label{sec:experiments}
We empirically evaluate each of our proposed methods separately. We demonstrate the usefulness of comparing contrastive explanations for explaining model adaptations in section~\ref{sec:experiments:explainingmodeldriftbydriftingcontrastiveexplanations}, and in section~\ref{sec:experiments:findingrelevantregionsindataspace} we evaluate our method for finding relevant regions in data space that are affected by the model adaptations and thus are interesting candidates for illustrating the corresponding difference in counterfactual explanations (see section~\ref{sec:finding_relevant_regions_in_dataspace}). Finally, we demonstrate the effectiveness of persistent local explanation for pushing the model adaptation towards a desired behaviour.

The Python implementation of all experiments is available on GitHub\footnote{\url{https://github.com/andreArtelt/ContrastiveExplanationsForModelAdaptations}}. We use the Python toolbox CEML~\cite{ceml} for computing counterfactual explanations and use MOSEK\footnote{We gratefully acknowledge a academic license provided by MOSEK ApS.} as a solver for all mathematical programs.

\subsection{Data Sets}\label{sec:experiments:datasets}
We use the following data sets in our experiments:
\paragraph{Gaussian Blobs Data Set}
This artificial toy data set consists of a binary classification problem and is generated by sampling from two different two dimensional Gaussian distributions - each class has is its own Gaussian distribution.
The drift is introduced by changing the Gaussian distributions between the two batches. In the first batch the two classes can be separated with a threshold on the first feature, whereas in the second batch the second feature must be also considered.

\paragraph{Boston Housing Data Set}
The ``Boston Housing Data Set''~\cite{BostonHousingDataSet} is a data set for predicting house-prices (regression) and contains $506$ samples each annotated with $13$ real and positive dimensional features.
We introduce drift by putting all samples with a \textit{NOX} value lower or equal than $0.5$ into the first batch and all other samples into the second batch.

\paragraph{Human Activity Recognition Data Set}
The human activity recognition (HAR) data set by~\cite{HarDataSet} contains data from $30$ volunteers performing activities like walking, walking downstairs and walking upstairs. Volunteers wear a smartphone recording the three-dimensional linear and angular acceleration sensors. We use a time window of length $64$ to aggregate the data stream and computed the median per sensor axis and time window. We only consider the activities \textit{walking}, \textit{walking upstairs} and \textit{walking downstairs}.
We create drift by putting half of all samples with label \textit{walking} or \textit{walking upstairs} into the first batch - i.e. the classifier has to distinguish walking vs. walking upstairs - and all other samples, the other half of \textit{walking} together with samples labeled as \textit{waking downstairs} into the other batch - i.e. for the second batch the classifier has to distinguish normal walking vs. walking downstairs.

\paragraph{German Credit Data Set}
The ``German Credit Data set''~\cite{GermanCreditDataSet} is a data set for loan approval and contains $1000$ samples each annotated with $20$ attributes ($7$ numerical and $13$ categorical) with a binary target value (``accept'' or ``reject''). We use only the first seven features: \textit{duration in month}, \textit{credit amount}, \textit{installment rate in percentage of disposable income}, \textit{present residence since}, \textit{age in years}, \textit{number of existing credits at this bank} and \textit{number of people being liable to provide maintenance for}.
We introduce drift by putting all samples where \textit{age in years} is less or equal than $35$ into the first batch and all other samples into the second batch.

\paragraph{Coffee}
\begin{figure}
  \caption{Labelwise mean spectra per measurement time.}
  \label{fig:datasets:coffee}
    \includegraphics[width=\textwidth]{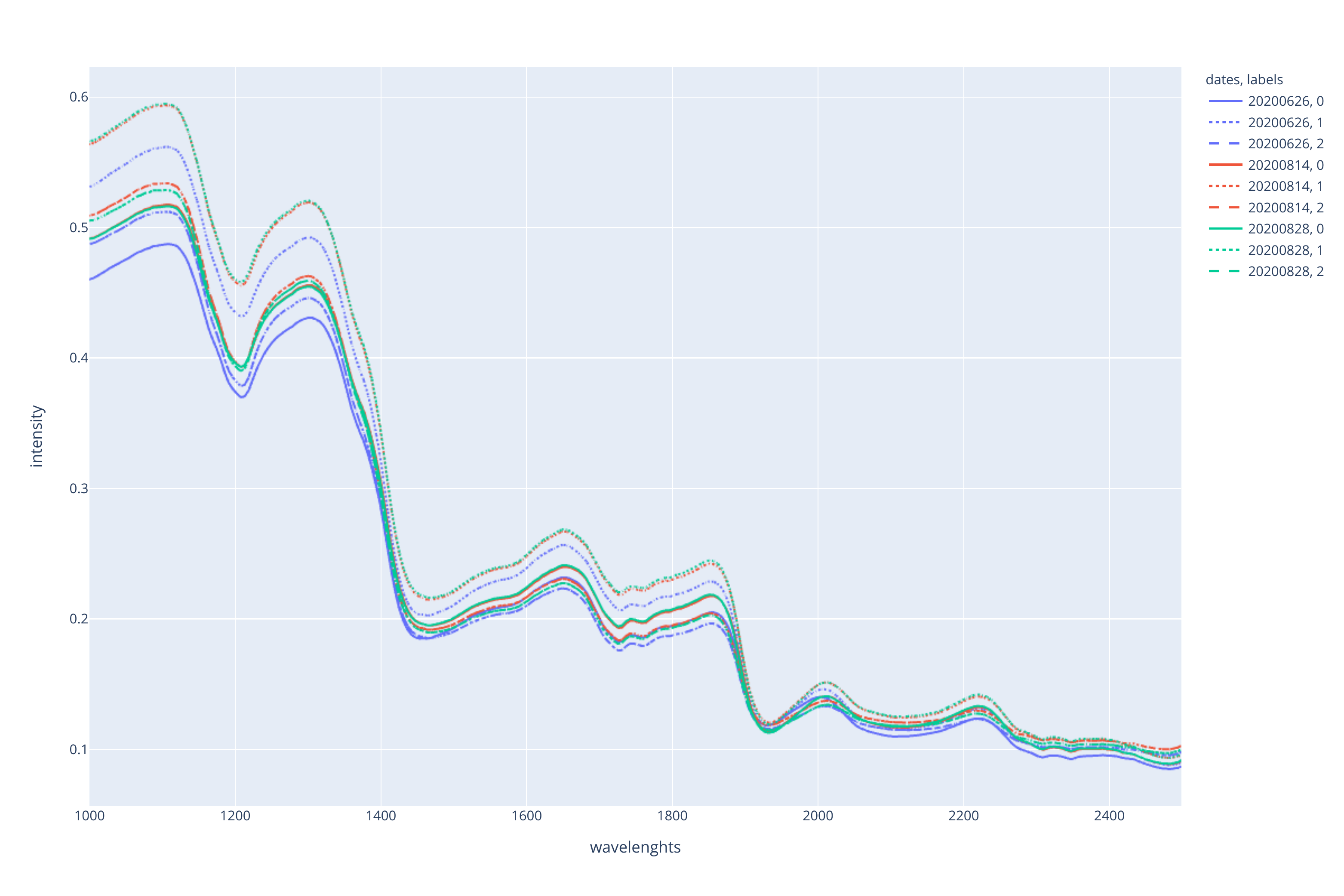}
\end{figure}
\FloatBarrier
The data set consists of hyperspectral measurements of three types of coffee beans measured at three distinct times within three month of 2020. Samples of Arabica, Robusta and immature Arabica beans were measured by a SWIR\_384 hyperspectral camera produced by Norsk Elektro Optikk. The sensor measures the reflectance of the samples for 288 wavelengths lying in the area between $900$ and $2500$nm. For our experiments, we standardize and subsample the data by a factor of $5$. Prior analysis of the data set indicates, that there the data distribution is drifting between the measurement times. Labelwise means of the data per measurement time are shown in Fig. \ref{fig:datasets:coffee}.

\subsection{Comparing Counterfactual Explanations for Explaining Model Adaptation}\label{sec:experiments:explainingmodeldriftbydriftingcontrastiveexplanations}
\subsubsection{Gaussian Blobs}
We fit a Gaussian Naive Bayes classifier to the first batch and then adapt the model to the second batch of the Gaussian blobs data set. Besides the both batches, we also generate $200$ samples (located between the two Gaussians) for explaining the model changes using counterfactual explanations.
We compute counterfactual explanations for all test samples under the old and the adapted model. The differences of the counterfactuals are shown in the right plot of Fig.~\ref{fig:exp:explainmodeldrift:gaussianblobs_houseprices}. We observe a significant change in the second feature of the adapted model - which makes sense since we know that, in contrast to the first batch, the second feature is necessary for discriminating the data in the second batch.

\subsubsection{Predicting House Prices}
\begin{figure}[!htb]
  \caption{Left: Changes in counterfactual explanations for the Gaussian blob data set. Right: Changes in counterfactual explanations for the house prices data set.}
  \label{fig:exp:explainmodeldrift:gaussianblobs_houseprices}
 \begin{minipage}[b]{0.49\textwidth}
    \includegraphics[width=\textwidth]{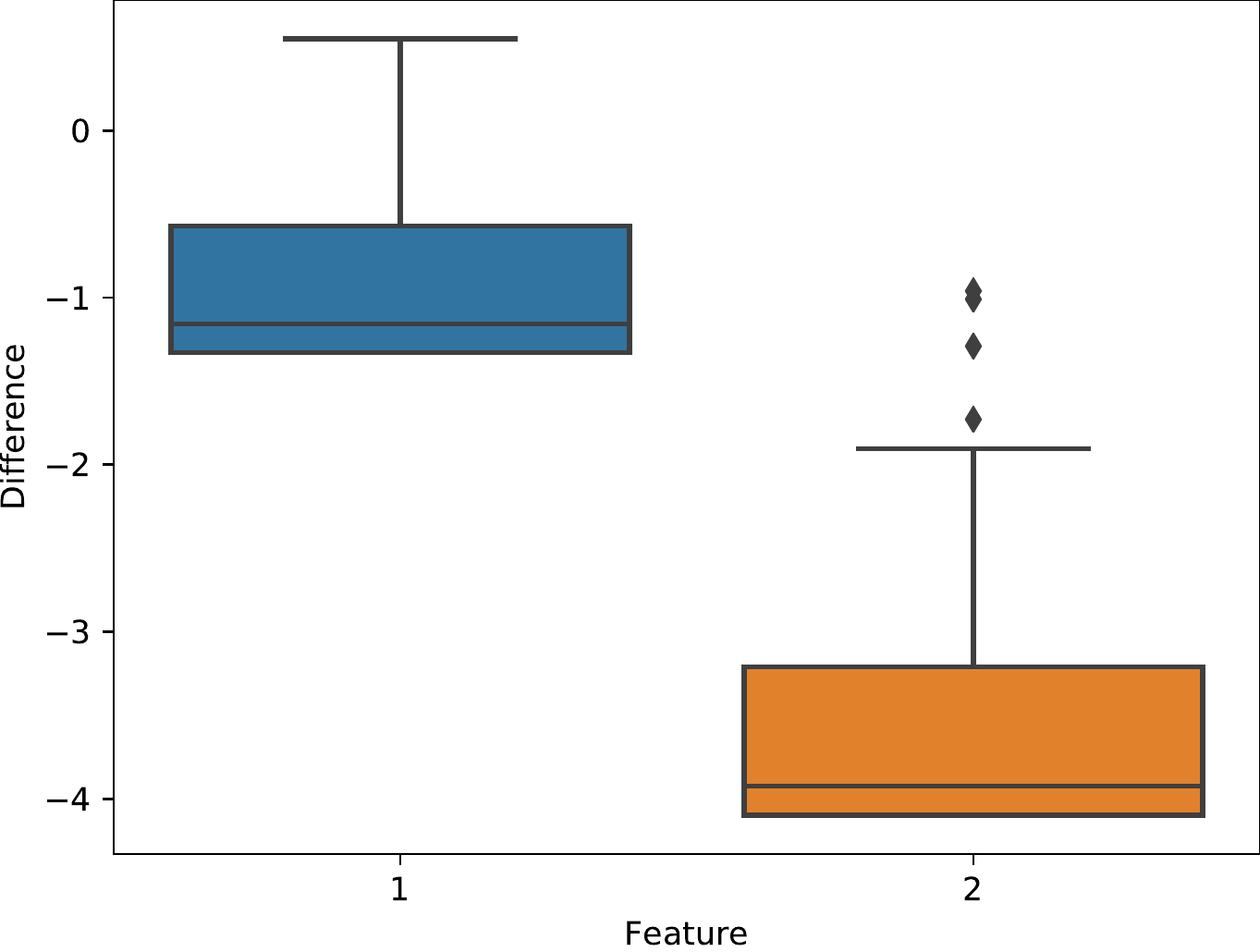}  
    \caption*{Gaussian blobs} 
   \end{minipage}
  \hfill
  \begin{minipage}[b]{0.49\textwidth}
    \includegraphics[width=\textwidth]{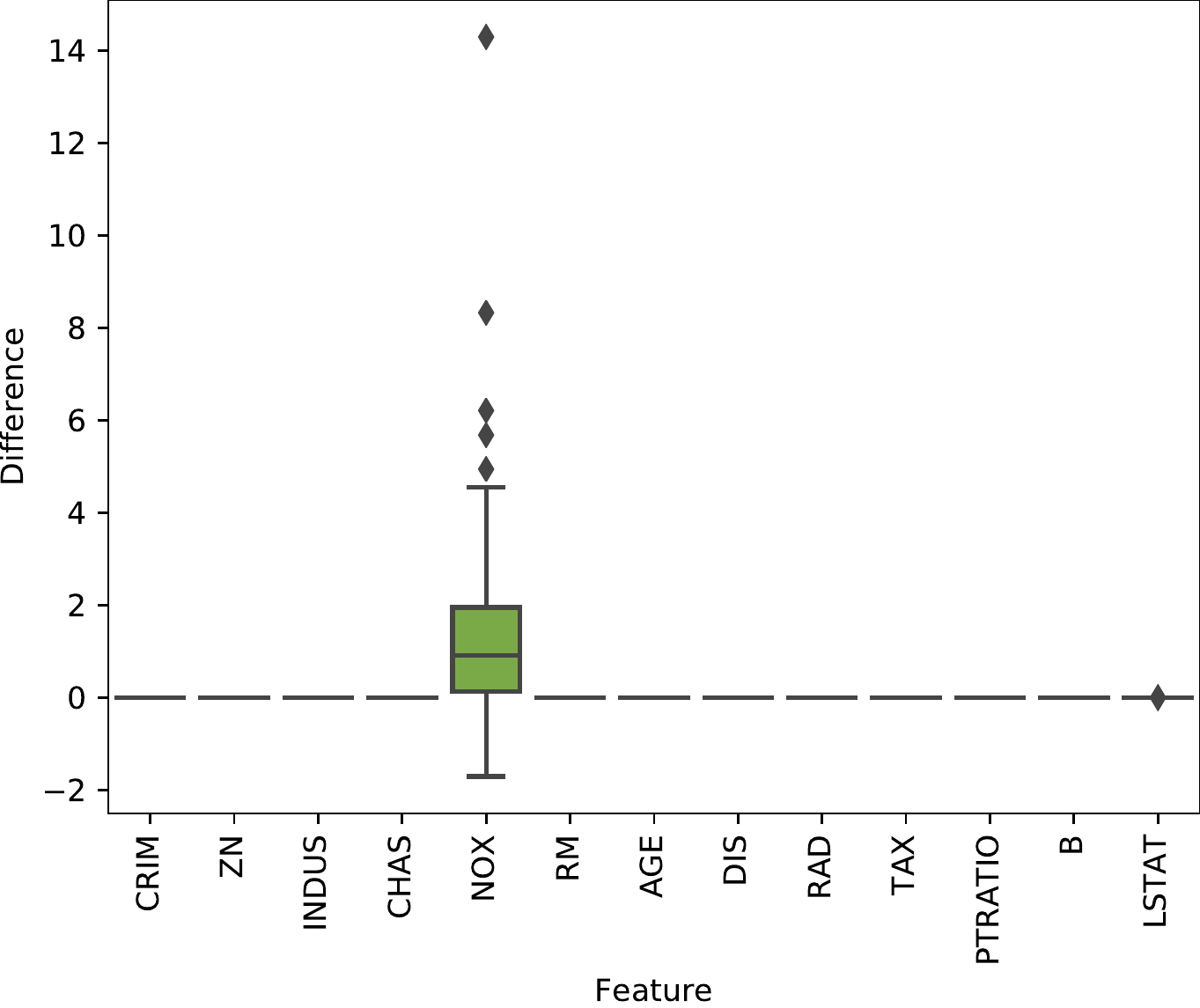}
    \caption*{House prices} 
  \end{minipage}
\end{figure}

We fit a linear regression model to the first batch and then completely refit the model on the first and second batch of the house prices data set.
We use the the test data from both batches for explaining the model changes using counterfactual explanations. We compute counterfactual explanations under the old and the adapted model whereas we always use a target prediction of $20$ and allow a deviation of $5$. The differences of the counterfactuals are show in the left plot of Fig.~\ref{fig:exp:explainmodeldrift:gaussianblobs_houseprices}. We observe that basically only the feature \textit{NOX} changes - which makes sense because we split the data into two batches based on this feature and we would also consider this feature to be relevant for predicting house prices.
\FloatBarrier

\subsubsection{Human Activity Recognition}
\begin{figure}
  \caption{Changes in counterfactual explanations - each target label is shown separately.}
  \label{fig:exp:explainmodeldrift:har}
 \begin{minipage}[b]{0.49\textwidth}
    \includegraphics[width=\textwidth]{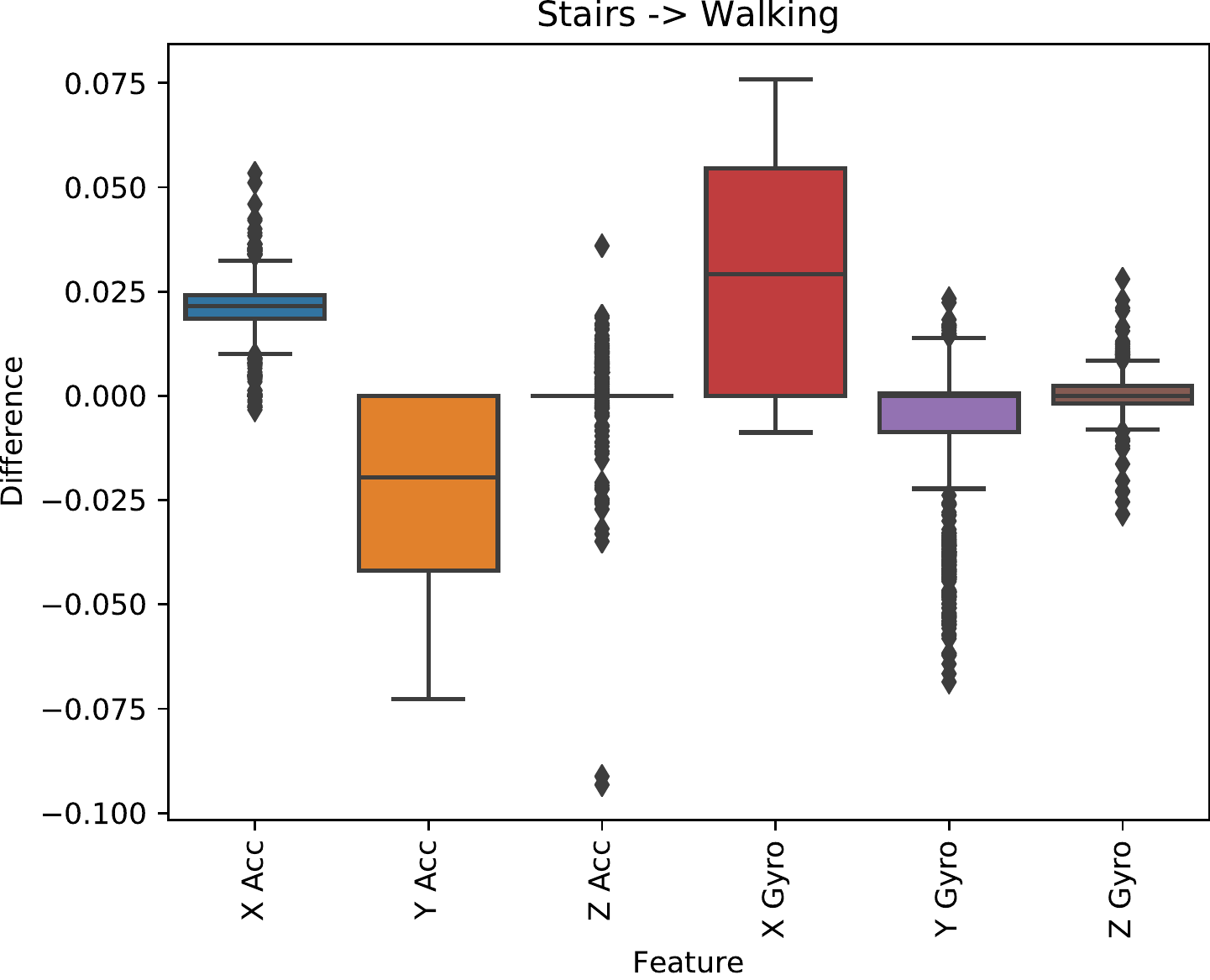}  
   \end{minipage}
  \hfill
  \begin{minipage}[b]{0.49\textwidth}
    \includegraphics[width=\textwidth]{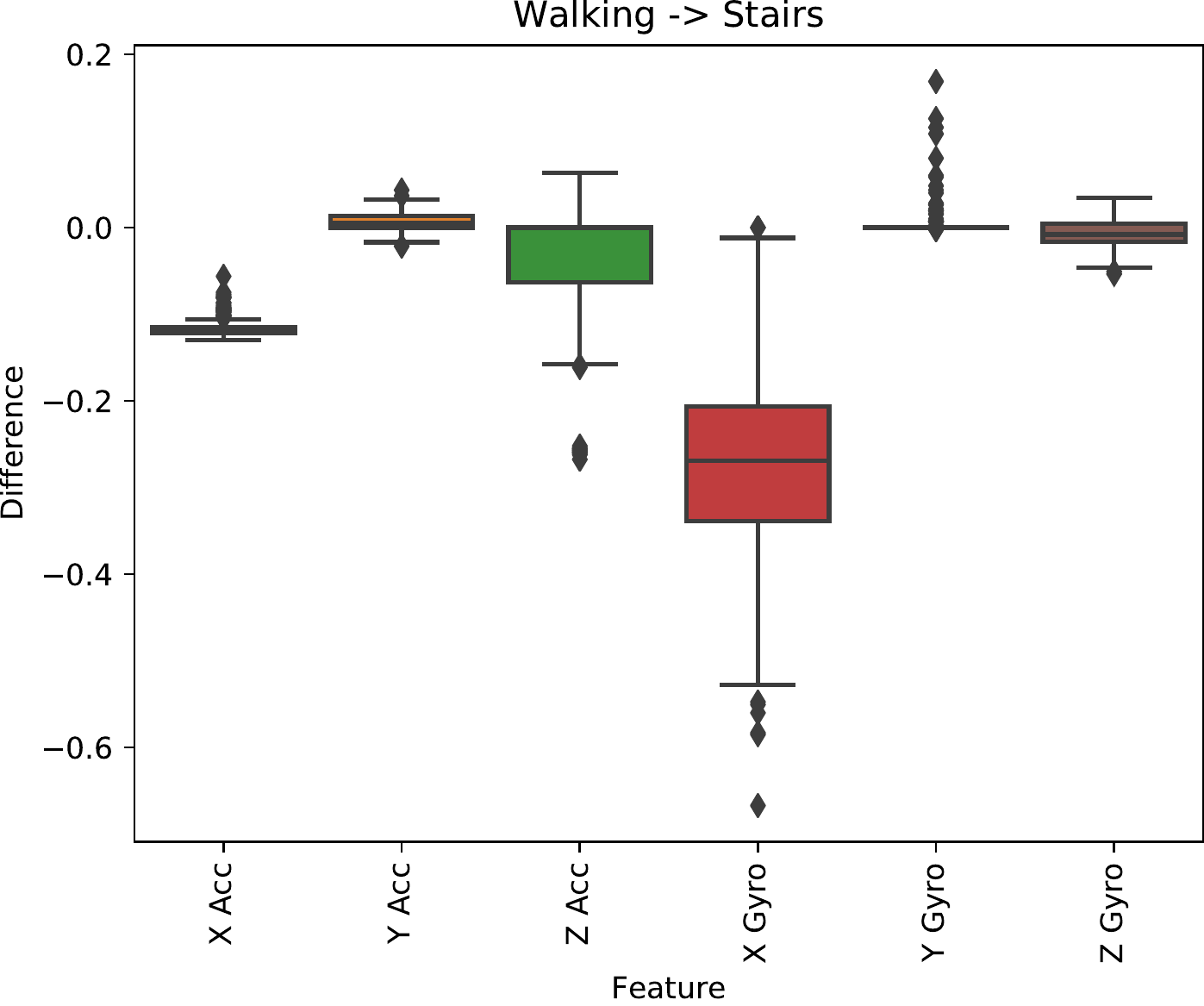}
 \end{minipage}
\end{figure}

We fit a Gaussian Naive Bayes classifier to the first batch and then adapt the model to the second batch of the human activity recognition data set. We use the the test data from both batches for explaining the model changes using counterfactual explanations. The differences of the counterfactuals (separated by the target label) are show in Fig.~\ref{fig:exp:explainmodeldrift:har}. In both cases we observe some noise but also a significant change in the Y axis of the acceleration sensor and the X axis of the gyroscope - both changes looks plausible because switching between walking up-/downstairs should affect the Y axis of the acceleration sensor while walking straight might be measurable by the X axis of the gyroscope, but since this is a real world data set, we do not really know the ground truth.
\FloatBarrier

\subsubsection{Coffee}
\begin{figure}
  \caption{Counterfactual explanations for two updated models.}
  \label{fig:exp:explainmodeldrift:coffee}
 \begin{minipage}[b]{0.49\textwidth}
    \includegraphics[width=\textwidth]{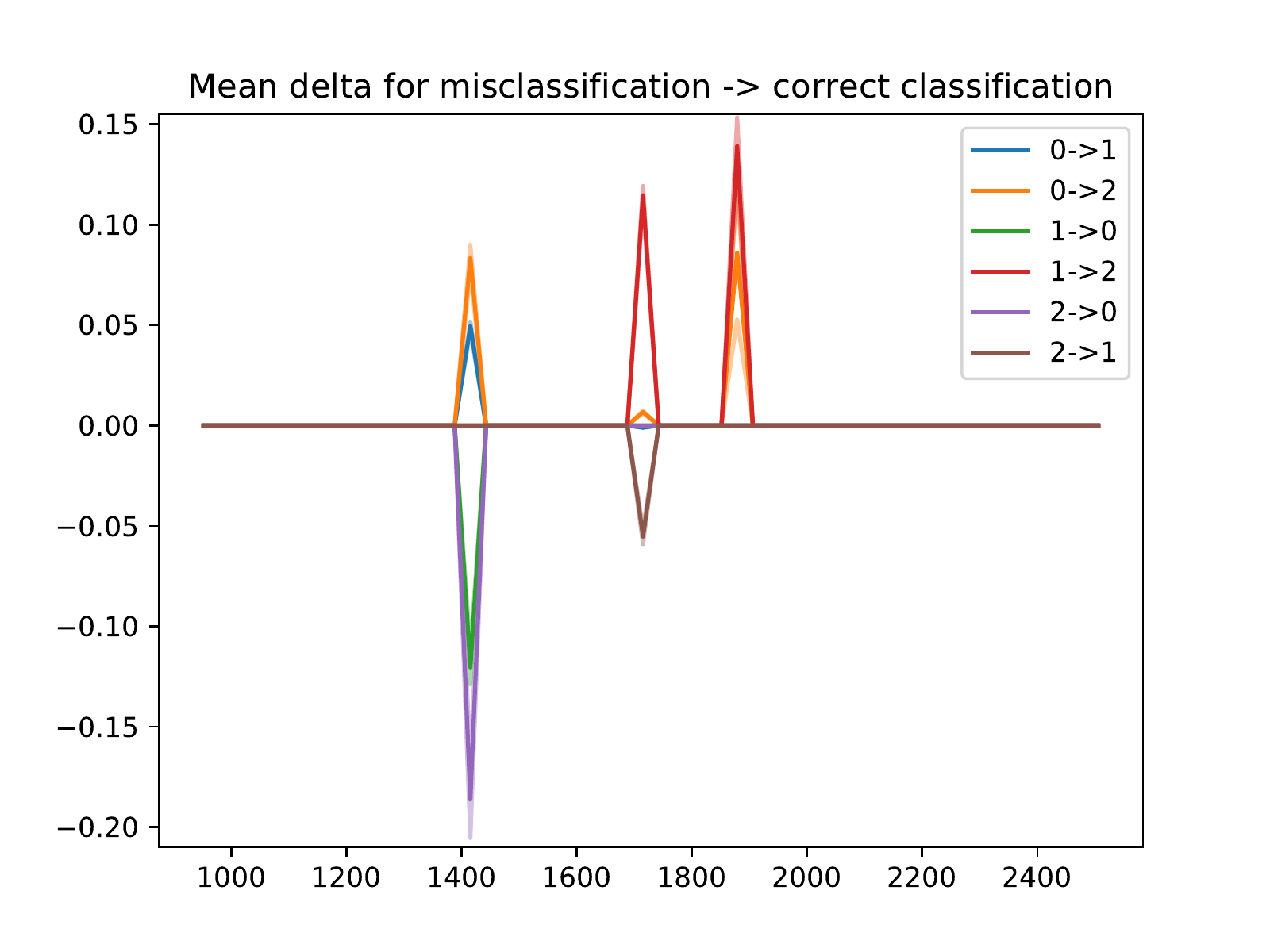}  
    \caption*{$t_2 =$ 14th August.} 
   \end{minipage}
  \hfill
  \begin{minipage}[b]{0.49\textwidth}
    \includegraphics[width=\textwidth]{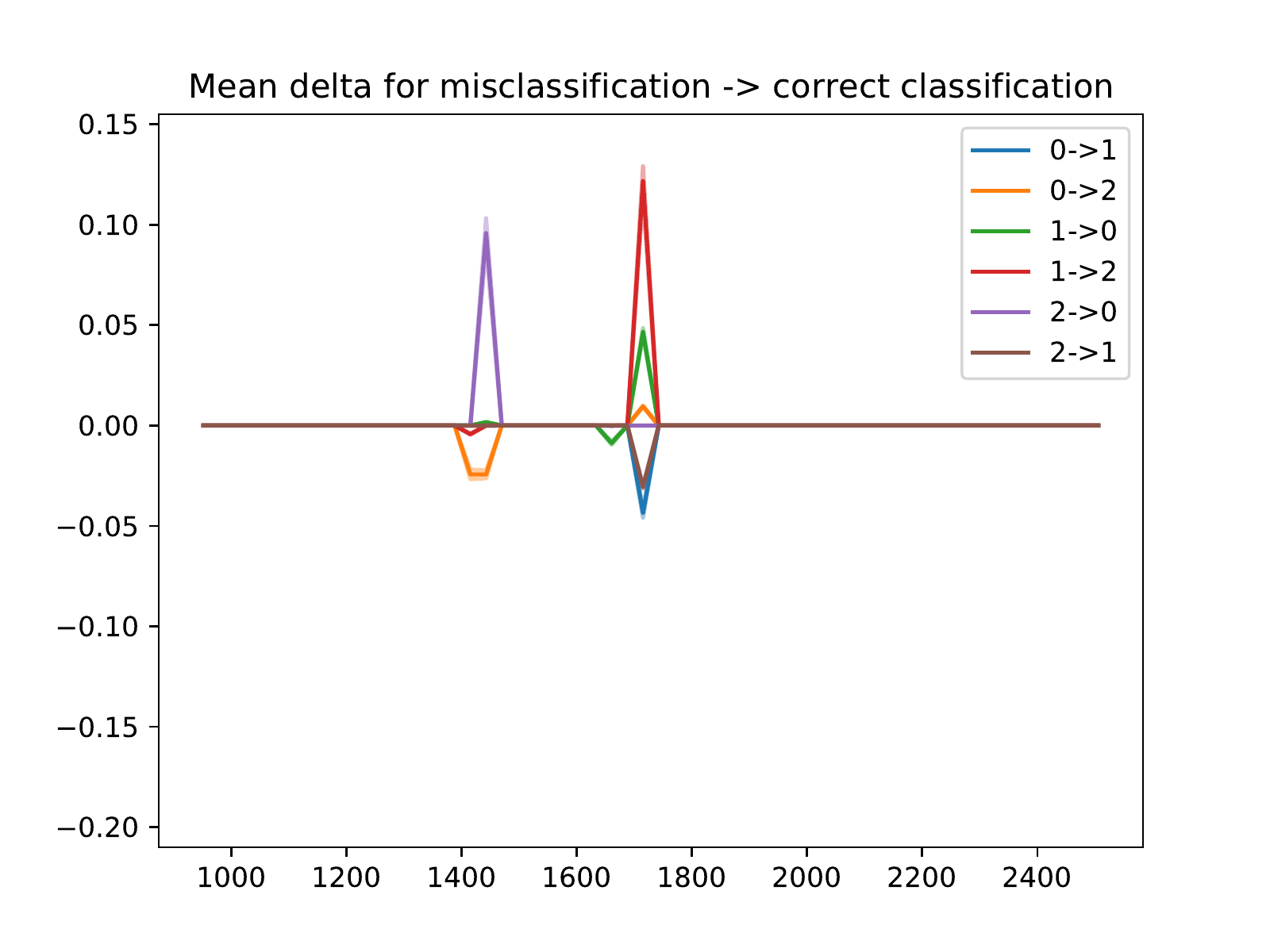}
    \caption*{$t_2 =$ on 28th August.} 
 \end{minipage}
\end{figure}

We are considering the model drift between a model trained with the data collected on the 26th June and another model based on the data from 14th August (from the 28th August in a second experiment). As the we know that the drift in our data set is abrupt, we train a logistic regression classifier on the training data collected at the first measurement time (model\_1), and another on the second measurement time (model\_2). We compute counterfactual explanations for all the samples in test set of the first measurement time that are classified correctly by model\_1 but misclassified by model\_2. The target label of the explanation is the original label. This way, we analyze how the model changes for the different measurement times. The mean difference between the counterfactual explanation and the original sample or visualized in Fig.~\ref{fig:exp:explainmodeldrift:coffee}. We observe that there are (interestingly) only a few frequencies which a are consistently differently treated by both model.
\FloatBarrier

\subsection{Finding Relevant Regions in Data Space}\label{sec:experiments:findingrelevantregionsindataspace}
\subsubsection{Gaussian Blobs}
\begin{figure}
  \caption{Left: Changes in counterfactual explanations considering all test samples. Right: Changes in counterfactual explanations considering the most relevant test samples.}
  \label{fig:exp:relevantsamples:gaussianblobs}
 \begin{minipage}[b]{0.49\textwidth}
    \includegraphics[width=\textwidth]{exp_results/gaussianblobs_all.pdf}  
   \end{minipage}
  \hfill
  \begin{minipage}[b]{0.49\textwidth}
    \includegraphics[width=\textwidth]{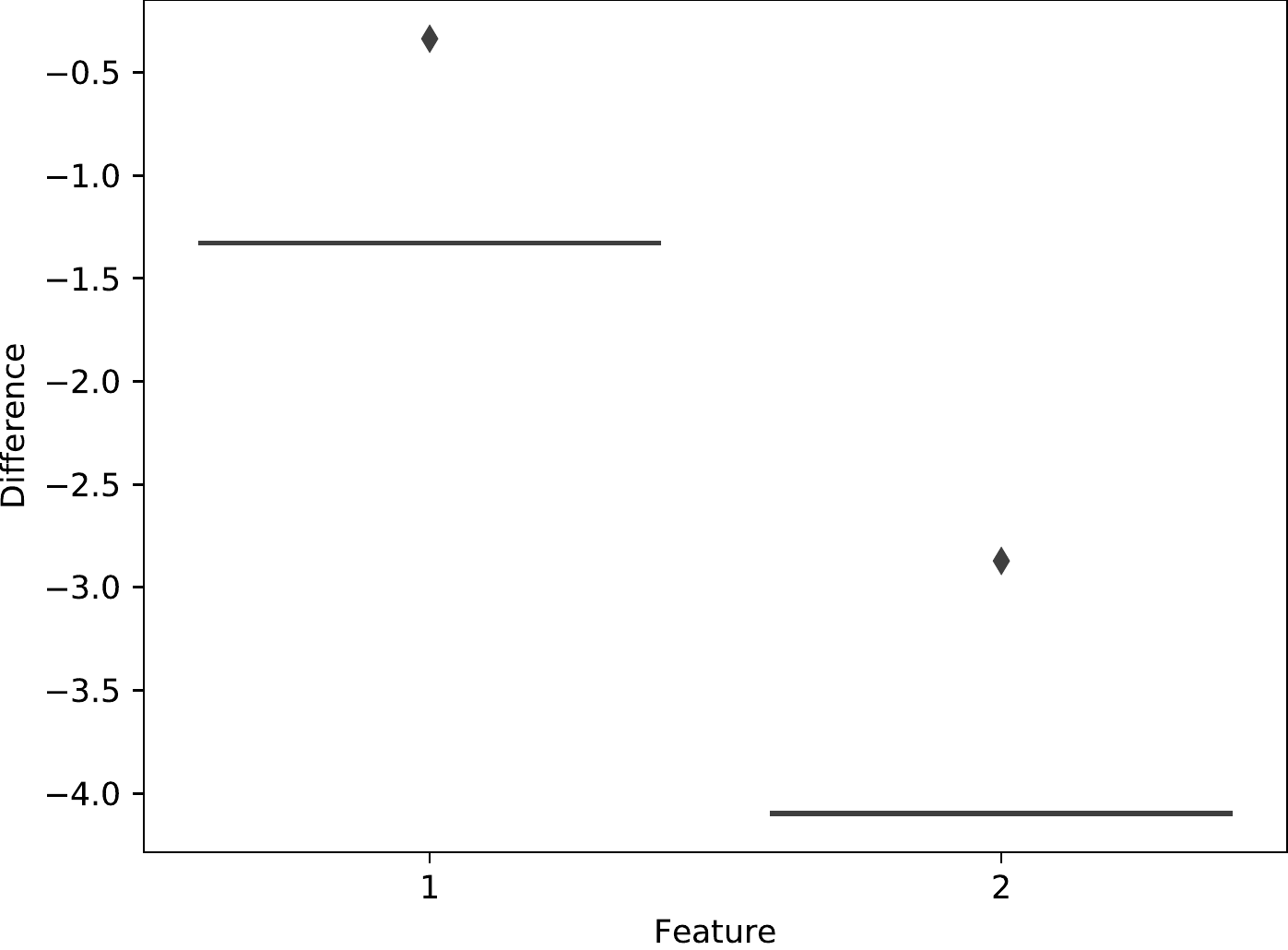}
 \end{minipage}
\end{figure}

We follow the same procedure like in section~\ref{sec:experiments:explainingmodeldriftbydriftingcontrastiveexplanations} but this time we do not use all test samples but only the $10$ (approximately $5$\% of the test samples) most relevant as determined by our method proposed in section~\ref{sec:finding_relevant_regions_in_dataspace}. In Fig.~\ref{fig:exp:relevantsamples:gaussianblobs} we plot the changes of the counterfactual explanations for both cases. We observe the same effects in both cases but with less noise in case of using only a few relevant samples - this suggests that our method from section~\ref{sec:finding_relevant_regions_in_dataspace} successfully identifies relevant samples for highlighting and explaining the specific model changes.
\FloatBarrier

\subsubsection{Human Activity Recognition}
\begin{figure}
  \caption{Left: Changes in counterfactual explanations considering all test samples. Right: Changes in counterfactual explanations considering the most relevant test samples.}
  \label{fig:exp:relevantsamples:har}
 \begin{minipage}[b]{0.49\textwidth}
    \includegraphics[width=\textwidth]{exp_results/har_stairs_walking_all.pdf}  
   \end{minipage}
  \hfill
  \begin{minipage}[b]{0.49\textwidth}
    \includegraphics[width=\textwidth]{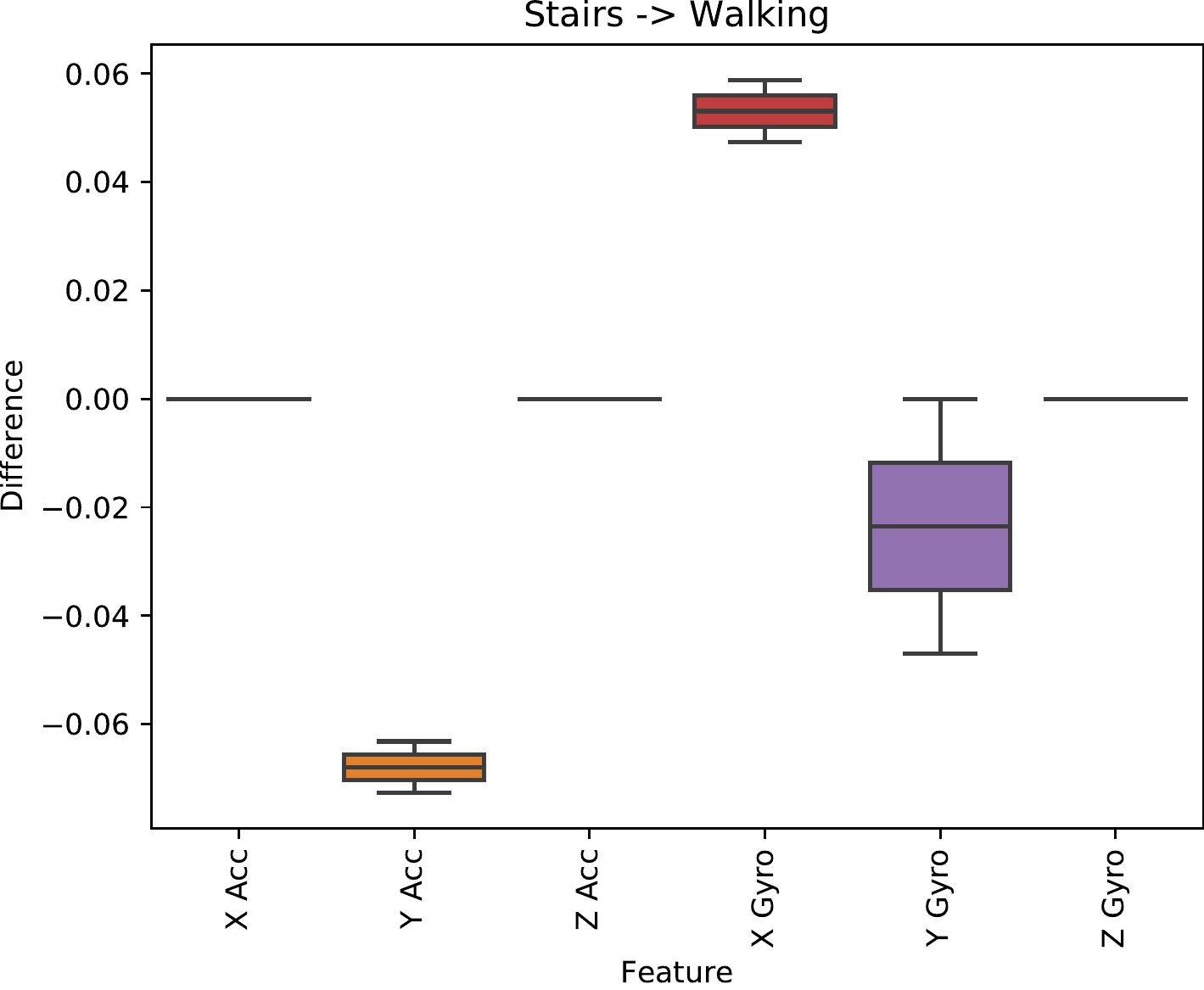}
 \end{minipage}
\end{figure}

We follow the same procedure like in section~\ref{sec:experiments:explainingmodeldriftbydriftingcontrastiveexplanations} but this time we do not use all test samples but only the $500$ (approximately $\frac{1}{3}$ of the test samples) most relevant as determined by our method proposed in section~\ref{sec:finding_relevant_regions_in_dataspace}. In Fig.~\ref{fig:exp:relevantsamples:har} we plot the changes of the counterfactual explanations for both cases when switching from walking up-/downstairs to walking straight. We observe the same effects in both cases but with much less noise in case of using only a few relevant samples (we also clearly observe a little change in the Y axis of gyroscope which is not that strong in case of using all samples) - this suggests that our method from section~\ref{sec:finding_relevant_regions_in_dataspace} successfully identifies relevant samples for highlighting and explaining the specific model changes.
Considering only the most relevant samples yields the same (but much stronger) results while saving a lot of computation time - this becomes even more handy when every sample has to be inspected manually (e.g. in some kind of manual quality assurance).
\FloatBarrier

\subsection{Persistent Local Explanations}\label{sec:experiments:persistentlocalexplanations}
\subsubsection{Loan approval}
\begin{figure}
  \caption{Left: Changes in counterfactual explanations. Right: Changes in counterfactual explanations under persistent counterfactual explanations.}
  \label{fig:exp:persistentexplanations:credit}
 \begin{minipage}[b]{0.49\textwidth}
    \includegraphics[width=\textwidth]{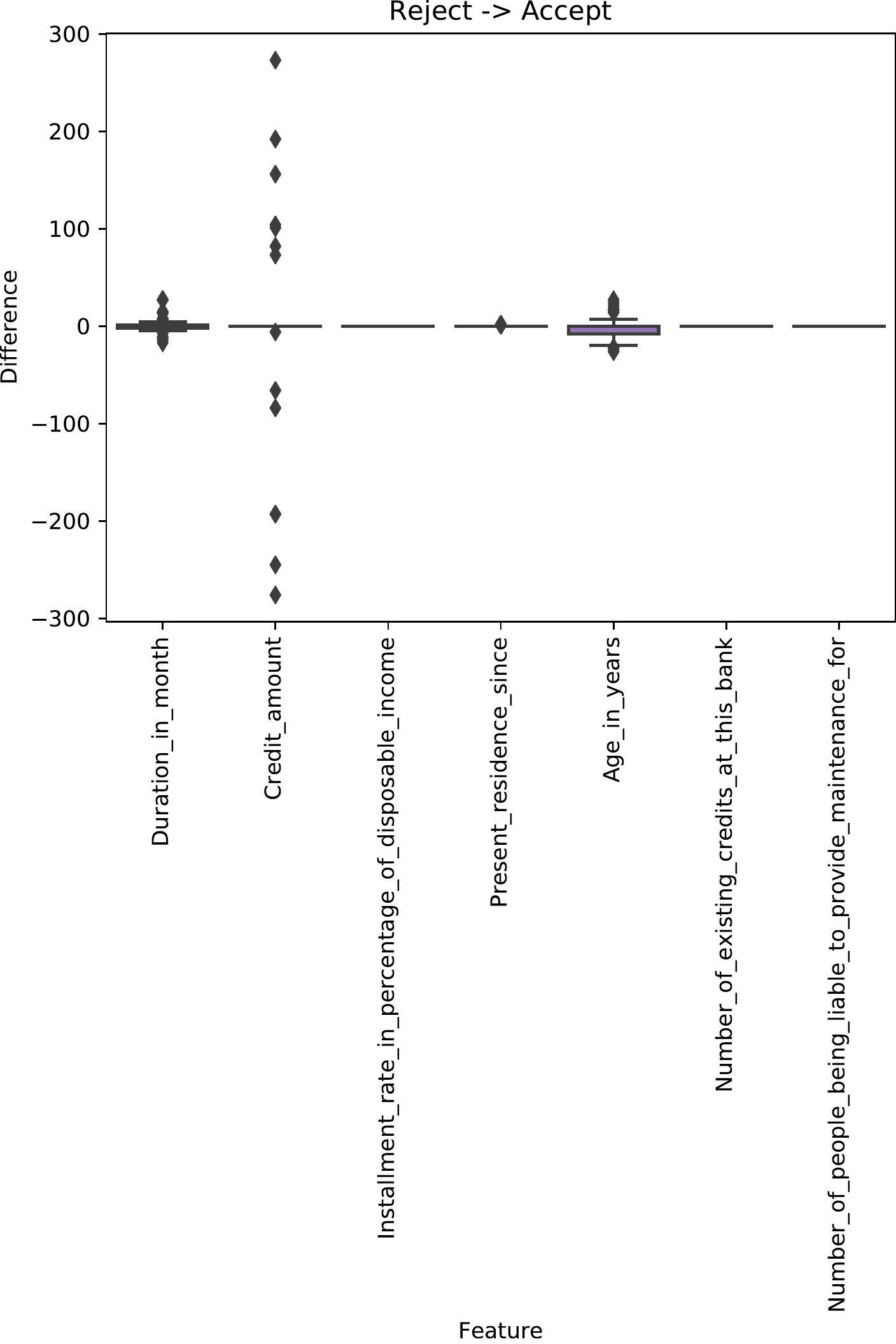}  
   \end{minipage}
  \hfill
  \begin{minipage}[b]{0.49\textwidth}
    \includegraphics[width=\textwidth]{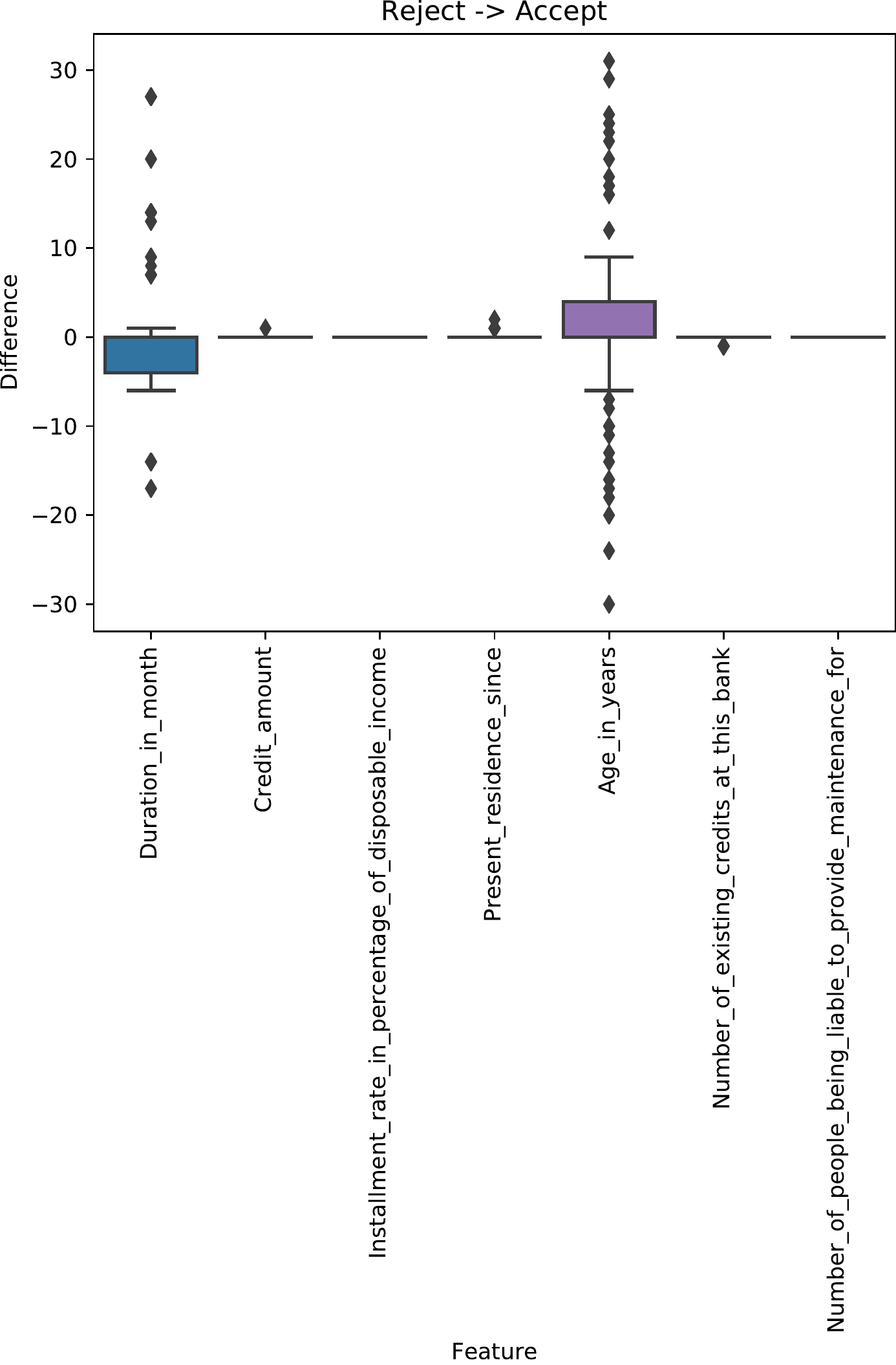}
 \end{minipage}
\end{figure}

We fit a decision tree classifier to the first batch and completely refit the model to the first and second batch of the credit data set. The test data from both batches is used for computing counterfactual explanations for explaining the model changes.
The changes in the counterfactual explanations for switching from ``reject'' to ``accept'' is shown in the left plot of Fig.~\ref{fig:exp:persistentexplanations:credit}. We observe that after adapting the model to the second batch (recall that we split data based on age), there are a couple of cases where increasing the credit amount would turn a rejection into an acceptance which we consider as inappropriate and unwanted behaviour. We therefore use our proposed method for persistent local explanations from section~\ref{sec:general:persitent_local_explanations} and section~\ref{sec:constrained_model_adaptations} to avoid this observed behaviour. The results of the constrained model adaptation is shown in the right plot of Fig.~\ref{fig:exp:persistentexplanations:credit}. We observe that now there is nearly no case in which increasing the credit amount turns a rejection into an acceptance - this suggests that our proposed method for persistent local explanations successfully pushed to the model towards our requested behaviour.


\section{Discussion and Conclusion}\label{sec:conclusion}
In this work we proposed to compare contrastive explanation as a proxy for explaining and understanding model adaptations - i.e. highlighting differences in the underlying decision making rules of the models. In this context, we also proposed a method for finding samples where the explanation changed significantly and thus might be illustrative for understanding the model adaptation. Finally, we proposed persistent constrative explanations for pushing the model adaptation towards a specific behaviour - i.e. ensuring that the model (after adaptation) satisfies some specified criteria.
We empirically demonstrated the functionality of all our proposed methods.

In future research we would like to study the benefits of comparing contrastive explanations for explaining model adaptations from a psychological perspective - i.e. conducting a user study to learn more on how people perceive model adaptations and how useful they find these explanations for understanding and assessing model adaptations. 


\bibliographystyle{splncs04}
\bibliography{bibliography.bib}

\appendix

\section{Proofs and Derivations}\label{sec:appendix}
\begin{itemize}
\item
\begin{proof}[\reftheorem{theorem:linearmodels:cosine}]
Given a sample $\x\in\RN^\dimsym$ and a correspond closest counterfactual $\xcf\in\RN^\dimsym$ (\refdef{def:counterfactual}) under a classifier $\classifier:\RN^\dimsym\to\setY$, we can compute the weight vector $\w\in\RN^\dimsym$ of a locally linear approximation of the classifier $\classifier(\cdot)$ between $\x$ and $\xcf$ as follows:
\begin{equation}\label{eq:locallinearapprox}
    \w = \xcf - \x
\end{equation}
Given a sample $\x\in\RN^\dimsym$ and a closest counterfactual $\xcf\in\RN^\dimsym$ before the model drift and another one $\xcfNew\in\RN^\dimsym$ after the model drift, we can compute the corresponding locally linear approximations of the decision boundaries~\refeq{eq:locallinearapprox} and compute the cosine angle between the two weight vectors~\refeq{eq:locallinearapprox} as follows:
\begin{equation}
\begin{split}
    \frac{\w_1^\top\w_2}{\pnorm{\w_1}_2\pnorm{\w_2}_2} &= \frac{\left(\xcf - \x\right)^\top\left(\xcfNew - \x\right)}{\pnorm{\xcf - \x}_2\pnorm{\xcfNew - \x}_2}\\
    &= \frac{\xcf^\top\xcfNew + \x^\top\x - \xcf^\top\x - \xcfNew^\top\x}{\sqrt{\left(\xcf^\top\xcf + \x^\top\x - 2\xcf^\top\x\right)\left(\xcfNew^\top\xcfNew + \x^\top\x - 2\xcfNew^\top\x\right)}}
\end{split}
\end{equation}
which concludes the proof.
\qed
\end{proof}

\item
\begin{proof}[\reftheorem{theorem:counterfactualdrift:linearmodel}]
The closest counterfactual $\xcf=\CF{\x}{\classifier}$ of a sample $\x$ under a linear binary classifier~\refeq{eq:linearclassifier} can be stated analytically~\cite{artelt2021fairness}:
\begin{equation}\label{eq:counterfactual:linearmodel}
    \xcf = \x - (\w^\top\x)\w
\end{equation}
Working out $\pnorm{\xcf - \xcfNew}_2^2$, where $\xcf=\CF{\x}{\classifierPrime}$, by making use of~\refeq{eq:counterfactual:linearmodel} and $\pnorm{w}_2=\pnorm{\wNew}_2=1$ yields:
\begin{equation}\label{eq:counterfactualdrift:linearmodel:dist}
\begin{split}
    \pnorm{\xcf - \xcfNew}_2 &= \pnorm{\x - (\w^\top\x)\w - \x + (\wNew^\top\x)\wNew}_2\\
    &= \pnorm{(\wNew^\top\x)\wNew - (\w^\top\x)\wNew + (\w^\top\x)\wNew - (\w^\top\x)\w}_2\\
    &= \pnorm{\left((\wNew - \w)^\top\x\right)\wNew + (\w^\top\x)(\wNew - \w)}_2
\end{split}
\end{equation}
Applying the triangle and Cauchy-Schwarz inequality to~\refeq{eq:counterfactualdrift:linearmodel:dist} yields:
\begin{equation}\label{eq:counterfactualdrift:linearmodel:dist:bound}
\begin{split}
    \pnorm{\xcf - \xcfNew}_2 &= \pnorm{\left((\wNew - \w)^\top\x\right)\wNew + (\w^\top\x)(\wNew - \w)}_2\\
    &\leq \pnorm{\left((\wNew - \w)^\top\x\right)\wNew}_2 + \pnorm{(\w^\top\x)(\wNew - \w)}_2\\
    &= \lvert(\wNew - \w)^\top\x\rvert_2\pnorm{\wNew}_2 + \lvert \w^\top\x \rvert_2 \pnorm{\wNew - \w}_2
\end{split}
\end{equation}
Applying the Cauchy-Schwarz inequality to~\refeq{eq:counterfactualdrift:linearmodel:dist:bound} yields:
\begin{equation}\label{eq:counterfactualdrift:linearmodel:dist:bound:part2}
\begin{split}
    \pnorm{\xcf - \xcfNew}_2 &\leq \lvert(\wNew - \w)^\top\x\rvert_2\pnorm{\wNew}_2 + \lvert \w^\top\x \rvert_2 \pnorm{\wNew - \w}_2\\
    &\leq \pnorm{\wNew - \w}_2\pnorm{\x}_2\pnorm{\wNew}_2 + \pnorm{\wNew}_2\pnorm{\x}_2\pnorm{\wNew - \w}_2\\
    &= 2\pnorm{x}_2\pnorm{\wNew - \w}_2
\end{split}
\end{equation}
Substituting $\pnorm{\wNew - \w}_2=\sqrt{2 - 2\cos(\angle\w,\wNew)}$ in~\refeq{eq:counterfactualdrift:linearmodel:dist:bound:part2} yields the stated bound:
\begin{equation}\label{eq:counterfactualdrift:linearmodel:dist:bound:part3}
\begin{split}
    \pnorm{\xcf - \xcfNew}_2 &\leq 2\pnorm{x}_2\pnorm{\wNew - \w}_2 \\
    &=2\pnorm{\x}_2\sqrt{2 - 2\cos(\angle\w,\wNew)}\\
    &=2\sqrt{2}\pnorm{\x}_2\big(1 - \cos(\angle\w,\wNew)\big)^{1/2}\\
    &=\sqrt{8}\pnorm{\x}_2\big(1 - \cos(\angle\w,\wNew)\big)^{1/2}
\end{split}
\end{equation}
which concludes the proof.
\qed
\end{proof}
\end{itemize}


%
\end{document}